\documentclass[10pt,journal,compsoc]{IEEEtran}
\usepackage{tikz}
\usetikzlibrary{bayesnet}

\usepackage{color}

\usepackage{xspace}
\usepackage[caption=false,font=footnotesize]{subfig}

\usepackage{algorithmic}

\usepackage{times}
\usepackage{epsfig}
\usepackage{pgfplots}
\pgfplotsset{compat=1.18}
\usepackage{pdfpages}
\usepackage{graphicx}
\usepackage{amsmath}
\usepackage{amssymb}
\usepackage{algorithm}
\usepackage{rotating}
\usepackage{xspace}
\usepackage[caption=false,font=footnotesize]{subfig}

\usepackage{setspace}
\usepackage{bm}

\newcommand{\comment}[1]{}
\def\etal{\emph{et al}\onedot}

\newcommand{\bigCI}{\mathrel{\text{\scalebox{1.07}{$\perp\mkern-10mu\perp$}}}}

\DeclareMathOperator*{\argmax}{\arg\!\max}

\definecolor{myBlue}{RGB}{189,235,255}
\definecolor{myGreen}{RGB}{235,255,189}
\definecolor{myRed}{RGB}{255,189,189}

\makeatletter
\DeclareRobustCommand\onedot{\futurelet\@let@token\@onedot}
\def\@onedot{\ifx\@let@token.\else.\null\fi\xspace}
\def\eg{e.g\onedot} 
\def\ie{i.e\onedot}

\def\etal{et al\onedot}
\makeatother

\ifCLASSOPTIONcompsoc
  \usepackage[nocompress]{cite}
\else
  \usepackage{cite}
\fi
\ifCLASSINFOpdf
\else
\fi
\begin{document}
%
\title{Person Identification from Contextual Motion}
%
%
%
%

\author{Igor~Kviatkovsky,
        ~Ehud~Rivlin,~\IEEEmembership{Senior Member,~IEEE}
				~and~Ilan~Shimshoni,~\IEEEmembership{Member,~IEEE}
\IEEEcompsocitemizethanks{
\IEEEcompsocthanksitem I. Kviatkovsky and E. Rivlin are with the Department of Computer Science, Technion -- Israel Institute of Technology, Technion City, Haifa 32000, Israel. \protect\\ E-mail: \{kviat,~ehudr\}@cs.technion.ac.il \protect\\
\IEEEcompsocthanksitem I. Shimshoni is with the Department of Information Systems, University of Haifa, Carmel Mount, Rabin building, Haifa 31905, Israel.	\protect\\	
E-mail: ishimshoni@mis.haifa.ac.il.

}
}

\IEEEtitleabstractindextext{%
\begin{abstract}
We consider the problem of identifying people based on their motion styles. We present a generative model describing the action instance creation process and derive a probabilistic identity inference scheme for two common person identification scenarios motivated by the surveillance and authentication applications. 
We introduce a novel, \emph{interactive}, scenario for person identification from motion patterns.
To this end, we formalize the identification process in the context of a sequential message exchange session between the subject and the system.
The subject's behavior is modeled using a probabilistic generative model inspired by the Human Information Processing (HIP) paradigm. 
At each stage, the system presents a visual stimulus (a cue) to the subject and records their motion response. 
The cue is selected so as to maximize the mutual information of the expected response and the subject's identity.
Once recorded, the response is used to update the a posteriori probability over possible subjects' identities.
The process terminates once a sufficient classification confidence level is reached. 
To the best of our knowledge, this is the first time person identification is addressed in such interactive setting.
We report high recognition rates on five publicly available datasets and our own novel dataset consisting of 4,476 recordings of 22 test subjects responding to 15 cues.
\end{abstract}

}

\maketitle

\section{Introduction}
\label{sec:intro}
Body language has been shown to be one of the most powerful features used by humans to infer identity and attributes such as age and gender~\cite{johansson,CuttingKozlowski,troje}. 
Human studies of body language have been the motivating force behind computer vision systems for automatic person identification from motion~\cite{trojePCA,WangGait,HanGait,hofmann,Munsell,Kviatkovsky_2015_CVPR_Workshops}. 
Most efforts focused on analyzing the discriminative properties of locomotion, \eg, gait, extracted from video sequences~\cite{WangGait,HanGait} or motion capture (mocap) 
data~\cite{trojePCA}. The growing popularity of depth camera sensors, \eg, Microsoft Kinect, has enabled additional modalities such as depth~\cite{hofmann} and skeleton~\cite{Munsell,Kviatkovsky_2015_CVPR_Workshops} to be used for this purpose. 

Person identification from motion rather than from more common alternatives such as face recognition is invaluable when the latter are unavailable or of insufficient quality.  
Such conditions may arise, for example, due to a limited camera field of view (FOV), challenging illumination conditions, an intentional disguise, or privacy considerations.
An interactive system with real-time 3D skeleton tracking capabilities may thus use motion patterns to identify individuals when their faces are either fully or partially outside its operational FOV or poorly illuminated. Since motion patterns are difficult to imitate in a natural and consistent manner, they may be used in surveillance or authentication systems to identify individuals carrying out fraud attempts when other biometric sources are either forged or intentionally disguised. 
On a different note, privacy policies prohibit installation of imaging devices in certain public facilities to prevent individuals' visual appearance information from being recorded. However, since skeleton motion sequences do not reveal the highly detailed visual description of an individual's appearance, but are nonetheless suitable for accurate individual identification, they are a realistic replacement for video as the primary visual modality, without compromising privacy.




\begin{figure}[tb]
\begin{center}
\begin{tabular}{c}
\includegraphics[width=8.5cm]{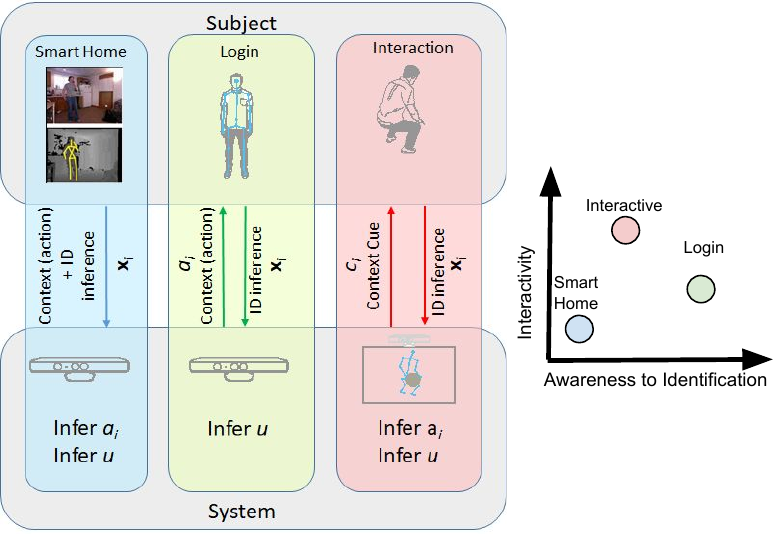} \\
~~~~~~~~~~~~~~(a)~~~~~~~~~~~~~~~~~~~~~~~~~~~~~~~~~~~~~~~~~~~(b)
\end{tabular}
\caption{
(a) Different scenarios for person identification from motion formalized as a communication between the subject and the system.
(b) Positioning of the scenarios on the interactivity vs. awareness scales.}
\label{fig:interactivity_awareness}
\end{center}
\end{figure}

In this work, we formalize the identification problem under three commonly encountered scenarios and present a unifying generative probabilistic framework as an effective solution. To be able to identify a person from a motion pattern, the pattern has to be reproducible, and therefore, definable. Motion patterns with a common semantic meaning are easily definable, \eg, ``walking'', ``throwing action''. Less common motion patterns, may be verbally described, \eg, ``extending the right arm to the left and moving it to the right''. In this work we use the term \emph{context} to refer to the type of action the person produced during the identification process.

Previous work treated person identification from motion under one of two scenarios. 
The first scenario assumes that subjects are interacting with the physical world and are not necessarily aware of the identification system's existence.
Person identification from gait~\cite{ChellappaGait,WangGait,HanGait,Egocentric,zhang2019gait} is one such example. Note that these works rely on a strong assumption about the 
context of the person's motion, which is walking. In our work we loosen this constraint by allowing person identification in a more general context, for example, 
identifying family members from their everyday actions using video streams from a home camera system. Unlike gait recognition, the system has to infer the 
context jointly with the family member identity.

Under the second scenario, both the subject and the system are fully cooperating to achieve the shared goal of identifying the subject. 
User authentication~\cite{BodyLogin, HandBiometrics,sae2019emerging}, where the subject is expected to perform a unique action to be granted access to the system, is one such example.
Note that this scenario is practical under the assumption that the authentication system explicitly sets the context for the subject. 

These two scenarios may be positioned on the so-called \emph{interactivity} scale, measuring the richness and the quality of the interaction between the subject and the system. The smart home scenario assumes zero interactivity, \ie, unilateral communication between the system and the subject. In the authentication scenario, the communication is bilateral, positioning it higher on the interactivity scale, but is limited to the bare minimum of information required by the authentication protocol. From the subject--system awareness point of view these two scenarios are positioned at the extremes (see Fig.~\ref{fig:interactivity_awareness}(b)).  

\begin{figure}[tb]
\begin{center}
\begin{tabular}{c}
\includegraphics[width=8.0cm]{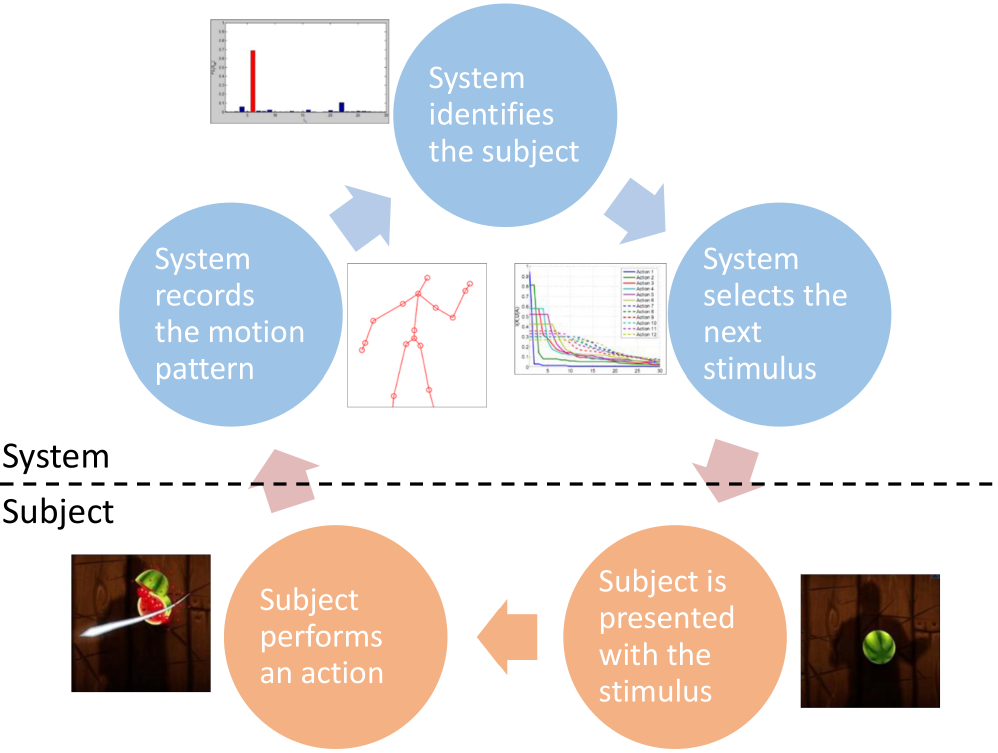}  
\end{tabular}
\caption{Interactive identification process. A cue is displayed, a person responds with an action which
is recorded and is used for identification. A new cue is generated so that the user’s
expected response maximizes its mutual information with respect to the identity.
}
\label{fig:process}
\end{center}
\end{figure}

In the present work we introduce a third scenario, positioned further on the interactivity scale and somewhere in between the extremes of the awareness scale (see Figs.~\ref{fig:interactivity_awareness}(b)).
Recent developments in the immersive natural user interface (NUI) offer a wide range of novel scenarios for person identification from motion. Under these scenarios, the subject interacts with the system; however, the goal of this interaction is not necessarily aligned with the goal of being identified, in contrast to the authentication scenario. Examples of such interactive activities include playing a Kinect game or manipulating applications through a gesture-based UI, where the subject is either partially or fully unaware of the underlying identification process. We show good results in identifying a person from arbitrary actions recorded during such interactive sessions. We assume that
the system--subject interaction follows an iterative (or sequential) process. 
In each iteration, the system generates a visual stimulus by displaying a moving object on the screen. The object's trajectory is predefined and is set to stimulate the subject to follow a certain dynamic pattern. 
The context here is set implicitly, turning the subject's experience into a more natural one. 
The subject's attempt to intercept the object, produces a body motion pattern that is recorded and used for identification. For the next iteration, the system decides on the stimulus that -- when presented to the subject -- is expected to reduce the identification uncertainty as much as possible. An example of such an interactive process is an extension of the Fruit Ninja game, where the fruit generation mechanism operates with two independent goals simultaneously: to provide users with the most enjoyable experience while successfully identifying them at the same time (see Fig.~\ref{fig:process}).

\subsection{Our Approach and Contributions}
\label{sec:approach}
The contribution of this work is threefold. First, we define three scenarios for person identification based on body motion and present a unifying probabilistic framework, 
based on generative models, for inferring the subject's identity under these scenarios. We show that, in spite of the relatively high noise levels common to low cost pose estimation devices, the individual motion patterns collected from everyday actions as a single cue for person identification, do indeed have good discriminative properties. 
Moreover, we show that the identification accuracy is significantly improved when a combination of various action types is used. 
We evaluate our approach on two publicly available datasets, originally created for evaluation of action recognition algorithms, in an experimental setting adapted for our needs. 

Second, out of the three defined scenarios, one is novel and assumes an interactive setting. The system--subject interaction is modeled as a communication over a noisy channel. In one direction, the system sends visual stimuli perceived by the subject, who is observing them on screen. In the opposite direction, the subject responds with motion patterns that the system receives through its imaging device. The exchanged messages are used to identify the subject. We define the identity of the currently observed subject as the system's true state and formulate the \emph{interactive identification} problem as a sequential process whose goal is state estimation. Since subjects do not
change during the interactive session, we may assume that the state is time independent. We use an active message selection framework based on the maximum mutual information (MMI) principle~\cite{DenzlerBrown,CoverThomas} to select the optimal messages to be sent to the subject at each iteration. The optimality criterion is set to reduce the state estimator's uncertainty. An interesting property of such a framework, proven in~\cite{DenzlerTR}, is that, given a particular choice of optimality criterion, the convergence of the process to the correct solution is theoretically guaranteed. 
Although our framework is based on~\cite{DenzlerBrown}, and is similar to~\cite{DarrellActiveGesture,GaoKoller,Karayev12timelyobject}, it is novel that the subjects themselves are actively manipulated. 
This is in contrast to other works, which actively select internal sensing parameters, e.g., camera viewpoint~\cite{DarrellActiveGesture,DenzlerBrown}, or apply post-processing
modules, e.g., sequences of feature extractors and classifier invocations~\cite{GaoKoller,Karayev12timelyobject}.

Finally, we present a fully functional instance of an interactive, real-time, person identification system, and introduce the CuedId dataset, a novel dataset recorded with the help of this system. We evaluate the performance of our approach using the CuedId dataset and a publicly available MSRC-12 dataset. We will make both the dataset and the code available for research purposes.

The rest of the paper is organized as follows. Section~\ref{sec:RelatedWork} discusses related work in the domain of motion-based person identification and active approaches in recognition. 
Section~\ref{sec:GenModel} formally defines the problem of person identification from motion using probabilistic generative models.
Section~\ref{sec:CuedId} focuses on the interactive scenario and introduces the Cued Person Identification (CPI) algorithm.
Section~\ref{sec:system} describes the experimental setup we have built for the evaluation of our approach.
In Section~\ref{sec:res}, the experimental results are presented, followed by a discussion and conclusions in Section~\ref{sec:disc}.

\section{Related Work}
\label{sec:RelatedWork}
\subsection{Action Recognition}
In the early works on action recognition, articulated pose
estimation was used to model the action, \eg,~\cite{yacoob}. However,
due to the high complexity of the articulated body
modeling, the attention has switched to appearance based
approaches~\cite{cuboids,LaptLind,ActDist}. 
Thanks to the recent developments in low cost depth sensors and the accompanying skeleton
tracking technologies~\cite{shotton}, model based approaches regained
their popularity~\cite{Actionlet,UTKinect,smij,Vemulapalli,Yao,HusseinCovIJCAI,Vemulapalli}.
With the introduction of deep learning, similarly to all other computer vision domains, 
action recognition schemes have switched to deep feature representations and deep sequence modeling~\cite{asadi2017deep}.      

The state-of-the-art in skeleton based action recognition has been advancing rapidly since the introduction of deep learning~\cite{shahroudy2016ntu,veeriah2015differential,liu2016spatio,zhu2016co,ke2017new,li2018independently}. 
This also would not be possible without the introduction of large datasets with skeleton sequences, such as the NTU RGB+D dataset~\cite{shahroudy2016ntu}. RNNs~\cite{veeriah2015differential} and LSTMs~\cite{liu2016spatio,zhu2016co,shahroudy2016ntu} have shown impressive performance in skeleton-based action recognition, however, are sometimes difficult to train due to the vanishing and exploding gradients, and are usually limited in the maximal possible sequence length. Independently Recurrent Neural Net (IndRNN)~\cite{li2018independently} is a recently introduced variation of RNN, which is much easier and faster to train, while resulting in state-of-the-art accuracy.

\subsection{Individual Recognition}
Person identification based on motion is a widely studied topic in the computer vision community~\cite{WangGait,HanGait,hofmann,Munsell,troje,Kviatkovsky_2015_CVPR_Workshops,HandBiometrics}.
Most efforts focus on analyzing locomotion, paying special attention to human
gait due to its applicability in the surveillance application
domain. Early works on gait-based recognition used
silhouette sequences extracted from video frames,~\eg,~\cite{WangGait,HanGait}.
Depth sensors have made other modalities, such
as depth~\cite{hofmann} and 3D skeletons~\cite{Munsell}, available for this task.
Munsell~\etal~\cite{Munsell} use sequences of 3D skeletal representations
from Kinect to model two types of locomotion -- walking and running. Given a test sequence, the authors
employ a two-step approach, first classifying the locomotion type and then applying a locomotion-specific
identity classifier.

Inspired by the seminal work on biological motion perception by Johansson~\cite{johansson}, many researchers investigated the option of classifying person identities and personal attributes from gait patterns, recorded using accurate mocap sensors attached to major body joints~\cite{trojePCA,troje,livne}. 
Troje~\cite{trojePCA} uses gait patterns to classify the walker's gender, concluding that it is better identified based on the gait dynamics than on the skeletal structure.  
The follow-up work by Troje~\etal~\cite{troje} extends the approach to identify additional personal attributes. 
Sigal~\etal~\cite{livne} automatically infer these attributes from videos of walking people using a 3D pose tracker.

In contrast to surveillance applications, where subjects are unaware of the underlying identification process, in the access-controlled authentication scenarios, subjects
voluntarily provide the system with their identification samples. Gesture-based biometrics is an evolving research field investigating, among other issues, whether human gestures may be used as an authentication modality~\cite{sae2019emerging}. 
Lai~\etal~\cite{Lai} demonstrate encouraging results using covariance descriptors extracted from silhouettes. In the follow-up works by Wu~\etal, the authors demonstrate improved results by replacing the silhouettes with Kinect skeletons~\cite{Wu} and explore the benefits of multiple viewpoints~\cite{BodyLogin}. 

The ``content-style'' generative model, introduced by
Tenenbaum and Freeman~\cite{TenFree}, represents an observation
as a bilinear mixture of content and style. This approach
was successfully applied in several domains including gait-based
person recognition~\cite{vasilescu,LeeElgammal}. The gait patterns were
decomposed into ``content'', \ie, the periodic gait pattern,
and ``style'', \ie, the personal stylistic variations, which was
used to recognize the person. 
However, the primary focus of these works was to automatically synthesize novel (never-seen-before) graphic animations, rather than to identify the actor.
Therefore, by decomposing action instances into content and style, they tried to simulate the actual action generation process.
In the conference version of this paper~\cite{Kviatkovsky_2015_CVPR_Workshops}, we addressed a simpler problem of discriminating between styles, skipping 
the discovery of underlying generative mechanism. We identified people using skeleton sequences of everyday actions, and showed that combining actions of 
various types significantly improves the classification accuracy. We present some of these findings in Section~\ref{sec:res}. 

Since the publication of~\cite{Kviatkovsky_2015_CVPR_Workshops}, there have been several interesting follow-up works. 
In~\cite{WuGestureStylesCNN} by Wu~\etal, a two-way CNN is shown to perform well in learning the user's gesture style from sequences of depth frames.
Chang and Park~\cite{ActionStylesRF} have presented a content-agnostic framework for style inference, based on Hough Forests~\cite{gall2011hough} of bag-of-pose features. The power of the approach is that it allows generalization to different content types beyond those used for training. 

Hoshen and Peleg~\cite{hoshen2016egocentric} have introduced a method for photographer identification from an egocentric video using a CNN. Recently, Zhang~\etal~\cite{zhang2019gait} have shown impressive results in gait recognition from video using a combination of an encoder-based CNN and an LSTM. To the best of our knowledge,~\cite{Wang&Wang} is the only work in the domain of person identification from everyday action skeleton motion sequences that has followed the deep learning trend. In~\cite{Wang&Wang} the authors have presented and evaluated several deep Recurrent Neural Net (RNN) architectures trained in a multi-task setting, jointly supervised by the action and subject labels. 

 


\subsection{Active Classification}
Active feature selection strategies are successfully employed in several areas of computer vision, \ie, object recognition~\cite{GaoKoller,Karayev12timelyobject,AnytimeReco,DenzlerBrown}, action recognition~\cite{DarrellActiveGesture}, object tracking~\cite{DenzlerTracking} and active sensing~\cite{DenzlerBrown}. In~\cite{GaoKoller}, a sequence of classifiers is selected to be applied on a given instance at test time. At each step, the classifier selection criterion is to maximize the expected information gain while simultaneously striving to minimize the incurred classification and feature extraction costs. In~\cite{Karayev12timelyobject,AnytimeReco}, the authors use reinforcement learning to learn instance-dependent, non-myopic strategies for selecting the best sequence of object detectors and scene classifiers to be applied on a given instance under budget constraints. The authors showed that this strategy outperformed random choice of object detectors and feature extractors. 
In~\cite{DenzlerBrown}, information theoretic considerations are used to select the best sensor parameters to improve object recognition and state estimation in general. 
In a follow-up work~\cite{DenzlerTracking}, the authors use these considerations to actively set the camera's focal length to improve object tracking accuracy. 
Our interactive identification framework relies on similar principles to those proposed in~\cite{DenzlerBrown}. 

Another closely related domain is \emph{foveated vision}. Darrel and Pentland~\cite{DarrellActiveGesture} use a reinforcement learning paradigm, based on Partially Observable Markov Decision Processes (POMDPs), to train a visual attention mechanism capable of foveating to acquire the most valuable information for classifying an individual's gesture. Although this work is similar to ours, it differs in several key aspects. First, we do not classify gestures, but use the motion data for identity classification. Second, and even more importantly, while in~\cite{DarrellActiveGesture}, and other related works~\cite{DenzlerBrown,DenzlerTracking}, the system's boundaries are defined so that the camera sensor is considered external, actively directed agent, in our setting, the human being is considered an actively manipulated external agent whose behavior is much less predictable and modellable.

\begin{figure}[tb]
\begin{center}
\begin{tabular}{cccc}
\begin{tikzpicture}
	\node[obs, fill=myBlue]             (x) {$\mathbf{x}_i$};
	\node[latent,  above=0.5cm of x]  (a) {$a_i$};
	\node[latent, right=0.3cm of x] (u) {$u$};

	\edge {u} {x} ; %
	\edge {a}{x};
	
	\plate {ax} {(x)(a)} {$N$} ;

\end{tikzpicture} &
\begin{tikzpicture}

	\node[obs, fill=myBlue]             (x) {$\mathbf{x}_i$};
	\node[obs, fill=myBlue, above=0.5cm of x]  (a) {$a_i$};
	\node[latent, right=0.3cm of x] (u) {$u$};

	\edge {u} {x} ; %
	\edge {a}{x};
	
	\plate {ax} {(x)(a)} {$N$} ;

\end{tikzpicture} &				
\begin{tikzpicture}

	\node[obs, fill=myBlue]             (x) {$\mathbf{x}_i$};
	\node[latent, above=0.5cm of x]  (a) {$a_i$};
	\node[obs, fill=myBlue, above=0.5cm of a]  (c) {$c_i$};
	\node[latent, right=0.3cm of x] (u) {$u$};

	\edge {u}{x} ;
	\edge {u}{a}%
	\edge {c}{a};
	\edge {a}{x};
	
	\plate {axc} {(x)(a)(c)} {$N$} ;

\end{tikzpicture} & 
\begin{tikzpicture}

	\node[latent]             (x) {$\mathbf{x}$};
	\node[latent, above=0.5cm of x]  (a) {$a$};
	\node[obs, fill=myBlue, above=0.5cm of a]  (c) {$c$};
	\node[latent, right=0.3cm of x] (u) {$u$};

	\edge {u}{x} ;
	\edge {u}{a}%
	\edge {c}{a};
	\edge {a}{x};

\end{tikzpicture}
\\

(a) & (b) & (c) & (d)
\end{tabular}
\caption{
Probabilistic generative models describing the action instance creation under different scenarios (Fig.~\ref{fig:interactivity_awareness}). 
(a) Smart home scenario, the context is inferred by the system.
(b) Login scenario, the context is set by the system.
(c) Interactive scenario, the context is implied by a visual cue displayed to the subject.
(d) A single iteration of the online active cue selection. Note that we marginalize over a yet to be obtained action instance.
}
\label{fig:GM}
\end{center}
\end{figure}
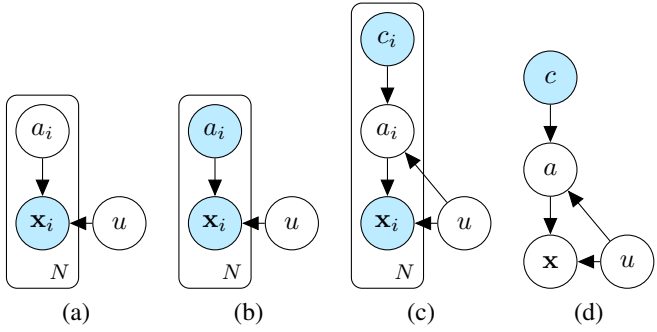

\section{Generative Models for User Identification}
\label{sec:GenModel}
In our model we restrict the allowed action types
to a set of atomic actions $\mathcal{A}$.
Let $\mathcal{U}$ denote the set of possible user
class labels. For each one of the scenarios presented in Fig.~\ref{fig:interactivity_awareness},
the action instances creation process is governed by one of the generative models in
Fig.~\ref{fig:GM}. Let $\{(\mathbf{x}_i,a_i)\}_{i=1}^N$ denote a set of $N$ pairs of random
variables, defined over $\mathbb{R}^d\times\mathcal{A}$, each associated with
$i$'s action instance representation and its label. Let $u$ denote
the random variable associated with the user identity
class, defined over $\mathcal{U}$. In the Login scenario (Fig.~\ref{fig:GM}(b)) we assume that action
labels are known, \ie, the variables $a_i$ are observed, while
in the Smart home scenario (Fig.~\ref{fig:GM}(a)) they are hidden. 

We will first consider the Login scenario since it is the most simple. 
Thus, $\{(\mathbf{x}_i,a_i)|\mathbf{x}_i\in\mathbb{R}^d,a_i\in\mathcal{A},i=1,\ldots,N\}$
denotes a set of $N$ action instances, performed by a certain user $u\in\mathcal{U}$.
According to the diagram in Fig.~\ref{fig:GM}(b),
\begin{equation}
\label{eq:post1}
p(u|\{(\mathbf{x}_i,a_i)\}_{i=1}^N)\propto p(u)\prod\limits_{i=1}^N p(\mathbf{x}_i|a_i,u),
\end{equation}
 
where $p(u)$ denotes a vector of prior probability values for all $u\in\mathcal{U}$ (see Appendix~\ref{appendix:ProbLogin} for details). 
The result $p(u|\{(\mathbf{x}_i, a_i)\}_{i=1}^N)$ is a vector of posterior probability values.

In the Smart home scenario we are given $N$ action instances $\{\mathbf{x}_i\}_{i=1}^N$
whose labels are unknown. By marginalization over $a_i$ we get,
\begin{equation}
\label{eq:post2}
p(u|\{\mathbf{x}_i\}_{i=1}^N)\propto p(u)\prod\limits_{i=1}^N \sum_{a_i\in\mathcal{A}}\frac{p(\mathbf{x}_i|a_i,u)}{p(\mathbf{x}_i|a_i)}p(a_i|\mathbf{x}_i),
\end{equation}

where $p(a_i|\mathbf{x}_i)$ is obtained by applying any given action
recognition algorithm on $\mathbf{x}_i$ (see Appendix~\ref{appendix:ProbSmartHome} for details). Note that Eq.~\ref{eq:post1} is a special
case of Eq.~\ref{eq:post2} assuming a perfect action recognition algorithm,
assigning 1 to $p(a_i|\mathbf{x}_i)$ if and only if $a_i$ is the $\mathbf{x}_i$'s true
label, and 0 otherwise. In such a case all the elements in the
sum vanish except for one, while the denominator $p(\mathbf{x}_i|a_i)$ is constant with respect to $u$.

In the interactive scenario described in Figure~\ref{fig:GM}(c), however, a cue $c$ is observed instead of the action label $a$.
In response, the user generates an action instance $\mathbf{x}$ of class $a$. 
Note that the Login scenario is a special case of the interactive one when each cue 
is deterministically translated into a unique action $a$. By marginalizing over possible response actions we get (see Appendix~\ref{appendix:ProbCuedId} for details),
\begin{equation}
\label{eq:post3}
p(u|\{(\mathbf{x}_i,c_i)\}_{i=1}^N) \propto p(u)\prod\limits_{i=1}^N \sum_{a_i\in\mathcal{A}} p(\mathbf{x}_i|a_i,u)p(a_i|c_i,u).
\end{equation}

\subsection{Classification}
\label{sec:classification}
We classify the user's identity using a MAP classifier corresponding to each one of Eq.~\ref{eq:post1}--~\ref{eq:post3}, depending on the scenario. For example, for the Smart home scenario (Eq.~\ref{eq:post2}):
\begin{equation}
\label{eq:map1}
u^*=\argmax_{u\in\mathcal{U}} p(u)\prod\limits_{i=1}^N \sum_{a_i\in\mathcal{A}}\frac{p(\mathbf{x}_i|a_i,u)}{p(\mathbf{x}_i|a_i)}p(a_i|\mathbf{x}_i).
\end{equation}

Assuming that we are given a set of labeled training samples, we use them to obtain a non-parametric estimate of
the likelihood distribution $p(\mathbf{x}|a,u)$ for all pairs $(a,u)\in\mathcal{A}\times\mathcal{U}$. Let $\mathcal{D}_{a,u}$ 
denote the set of action instances of user $u$ performing action $a$. Thus, applying a 1-nearest neighbor kernel density estimation (KDE), 
we obtain an estimator $\widehat{p}(\mathbf{x}|a,u)$ for $p(\mathbf{x}|a,u)$:
\begin{equation}
\label{eq:kde}
\widehat{p}(\mathbf{x}|a,u)\equiv \frac{1}{|\mathcal{D}_{a,u}|V(r)},
\end{equation}

where $V(r)$ is the volume of the $D$-dimensional\footnote{$D$ is the intrinsic data dimension, \ie, the dimension of the actual manifold in which the data resides.} 
sphere of radius \[r=\min_{\mathbf{x}'\in\mathcal{D}_{a,u}\setminus\{\mathbf{x}\}}d_a(\mathbf{x},\mathbf{x'}),\] centered at $\mathbf{x}$, and 
$d_a(\cdot,\cdot)$ measures the distance between action instances of class $a$.
We show in Section~\ref{sec:distMetric}, that the use of an action-specific distance measure allows us to take into account the instance variability of each particular action,
when discriminating between users.

\subsection{Action Representation}
\label{sec:rep}
There exists a wide range of approaches for representing
skeletal joint ensembles in 3D. Joint positions~\cite{HusseinCovIJCAI}, normalized
joint positions~\cite{UTKinect}, joint angles~\cite{smij}, pairwise relative
positions~\cite{Actionlet}, geometric boolean features~\cite{Muller,Yao} and
points in a Lie group~\cite{Vemulapalli} -- all of these proved useful for action
recognition and computer animation. In this work we
adopt one of the simplest existing skeletal representations,
namely the normalized joint positions (JP), due to its recent
success in action recognition reported in~\cite{Vemulapalli}. The JP representation
is constructed by taking the absolute joint positions
in 3D and normalizing them with respect to the hip
joint position. In the case of Kinect, the resulting representation
is a $(19\times 3)$-dimensional vector.

We represent each action instance as a concatenation of the JP representations of its poses.
To be able to represent the actions in $\mathbb{R}^d$, it is necessary to normalize the number of poses in each instance. To this end, we use cubic spline interpolation to temporally normalize all instances to an equal number of $L$ poses. We experimented with several values of $L$ and found that, for actions with a well defined semantic meaning, used to evaluate the Login and Smart home scenarios, $L=60$ achieves the best tradeoff between accuracy and computational efficiency. For more primitive actions, such as those used in the Interactive scenario evaluation, we set $L=15$. 

\subsubsection{Temporal Normalization}
\label{sec:TempNorm}
Rate variation is a well known problem in comparing action
instances performed under different conditions (speed,
style, etc.). Several methods were proposed to represent the
action in a rate-invariant fashion. Veeraraghavan~\etal~\cite{nominalCurve}
use a variant of Dynamic Time Warping (DTW)~\cite{MullerDTW} to handle
this issue. Wang~\etal~\cite{Actionlet} build a Fourier Temporal
Pyramid (FTP) and represent the action using the low frequency
coefficients. In~\cite{Vemulapalli},
the authors combine the DTW and the FTP to
obtain state-of-the-art performance in action recognition.
While it is intuitive why temporal normalization helps action
recognition it is not so with person identification. The
difference in rate may be exactly what differentiates between
users. In practice we saw that temporal normalization
is indeed beneficial for user identification as well.

\subsection{Action-Specific Distance Metric Learning}
\label{sec:distMetric}
\begin{figure}[ht]
\begin{center}
\begin{tabular}{cc}
\includegraphics[width=4cm]{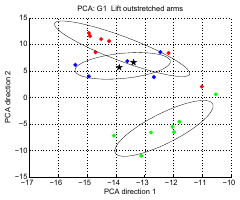} & 
\includegraphics[width=4cm]{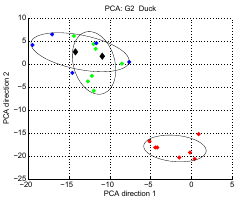} \\ 
(a) & (b)
\end{tabular}
\caption{
Instance representations of the ``Lift outstretched arms''
(a) and ``Duck'' (b) actions from the MSRC-12~\cite{MSRC12}
dataset performed by three users. The representations are projected
on the two major components of the PCA subspace.
}
\label{actionsPCA}
\end{center}
\end{figure}

Due to the highly constrained nature of the human skeleton
structure and its dynamics~\cite{troje,yacoob}, most of the skeletal
motion representations are redundant. Therefore, we reduce
the instance dimensionality for each action class $a\in \mathcal{A}$
using Principal Component Analysis (PCA). The resulting
representation is compact and less noisy, but considering
our final goal we are interested in transformations improving
the discriminative capabilities of the identity classifier.
Therefore, we apply Linear Discriminant Analysis (LDA)
on the PCA-transformed representations. The LDA maximizes
the ratio of the between-class scatter to the within-class
scatter so that the transformed instances of different
identities fall far apart while those belonging to the same
identity fall closer to each other. After applying LDA on
samples shown at Fig.~\ref{actionsPCA}, the overlapping regions will be reduced
while the clusters will be contracted as much as possible.

To conclude, let $D_p$ and $D_l$ denote the number of components selected in PCA and LDA, respectively. Given the
PCA and the LDA transformations, $\mathbf{P}_a^{D_p\times D}$ and $\mathbf{L}_a^{D_l\times D_p}$ obtained from the $D$-dimensional instances of 
action $a\in\mathcal{A}$, we define $\mathbf{Z}_a^{D_l\times D}=\mathbf{L}_a\mathbf{P}_a$ as the composite PCA-LDA
transformation for action class $a$.

Given a set of learned dimensionality reduction transformations,
$\{\mathbf{Z}_a|a\in\mathcal{A}\}$, we define the distance between a
pair of action instance representations $\mathbf{x}_i$ and $\mathbf{x}_j$ of class $a$,
as a squared Euclidean distance between their low dimensional
representations. Thus,
\begin{align*}
d_a(\mathbf{x}_i,\mathbf{x}_j)&=\left\|\mathbf{Z}_a\mathbf{x}_i-\mathbf{Z}_a\mathbf{x}_j \right\|_2^2=(\mathbf{x}_i-\mathbf{x}_j)^T\mathbf{M}_a(\mathbf{x}_i-\mathbf{x}_j)\\
&=\left\|\mathbf{x}_i-\mathbf{x}_j\right\|_{\mathbf{M}_a},
\end{align*}

where $\left\|\cdot\right\|_{\mathbf{M}_a}$
is a Mahalanobis distance and $\mathbf{M}_a = \mathbf{Z}_a^T\mathbf{Z}_a$.
We use this action-specific distance measure for computing
the likelihood estimator in Eq.~\ref{eq:kde}. Note that the number of
action instances in the training set for each action-user pair,
$\mathcal{D}_{a,u}$, has to be larger than three in order for the PCA-LDA to
work properly.

\subsection{Action Recognition for User Identification}
\label{sec:ActReco}
The MAP identity classifier in Eq.~\ref{eq:map1} depends on action
recognition performance via the term $p(a|\mathbf{x})$. In our
experiments we use one of the action classifiers proposed
in~\cite{Vemulapalli}, namely the linear SVM classifier trained in a 
one-vs-all fashion using normalized joint position (JP) representation
(see~\cite{Vemulapalli} for details). We refer to this classifier
as to JP-SVM. A given action instance, is classified as belonging
to the class with the largest JP-SVM classification
margin. We assign probability 1 to this class. An alternative
approach of assigning each class a probability proportional to its
margin, resulted in poor performance.

In our experiments we use the same datasets to train the
user and the action classifiers. Thus, special care should
be taken to ensures that, at test time, action classifiers are
applied only on action instances from different users than
those used during training. We solve this by using a 
leave-one-out approach to train the JP-SVM classifiers. That is,
we train $N_u$ action classifiers so that classifier number $u$
is trained on action instances from all users except $u$. At
test time, when computing the posterior $p(u|\{\mathbf{x}_i\}_{i=1}^N)$ we
always use the action classifier which was not exposed to
$u$'s instances during training.

\subsection{Homogeneous and Heterogeneous Instance Sets}
\label{sec:HomoVsHeteroExplain}
The MAP classifier given in Eq.~\ref{eq:map1} classifies the user
based on a set of instances, $\{\mathbf{x}_i\}_{i=1}^N$, belonging to certain
action types, regardless of whether the actual action labels
are given or not. We categorize the sets, of size larger than
one, into two categories, namely the \emph{homogeneous} and \emph{heterogeneous}
sets. A homogeneous set contains identically
labeled action instances, \ie, $\forall_{i,j},a_i=a_j$, while all instances
in a heterogeneous set are labeled differently, \ie, $\forall_{i\neq j},a_i\neq a_j$.
Figs.~\ref{actionsPCA}(a) and~\ref{actionsPCA}(b) show the distributions of
the ``Lift outstretched arms'' and the ``Duck'' actions, from
the MSRC-12~\cite{MSRC12} dataset, respectively. The instances come
from three users, marked with red, blue and green circles.
The chosen actions utilize the torso and the legs differently,
reflecting the various motion modalities of the human body.
Note that the ``red'' user performs the first action similarly to
the ``blue'' one and differently from the ``green'' one, while
the ``green'' user performs the second action similarly to the
``blue'' one and differently from the ``red'' one. The black
``$\star$''s and ``$\Diamond$''s are the action instances used to identify the
user. Using a homogeneous set of two stars for this task will
result in high probability values for the ``blue'' and ``red''
users. A similar ambiguous result -- now for the ``blue'' and
``green'' users -- is expected if a homogeneous set of two diamonds
is used. On the other hand, using a heterogeneous
set of one star and one diamond will unambiguously identify
the ``blue'' user. We remark that an alternative solution
of increasing the homogeneous set size, $N$, is not feasible
since $N$ does not scale well with an increasing number of
users and an increased area of the overlapping regions belonging to different 
users. We present extensive experimental validation
of the above-mentioned claims in Section~\ref{sec:HomoVsHetero}.

\section{Cued Person Identification}
\label{sec:CuedId}
The process of interactive person identification, informally described in Section~\ref{sec:intro}, assumes an interactive \emph{session} between a subject and the system, in which the parties exchange visual messages not necessarily related to the identification process itself. The goal of the system is to identify the subject as quickly as possible based on his or her motion in response to the generated \emph{cues} (visual stimuli). 

We make several simplifying assumptions as to the system--subject interaction, to facilitate a formal definition and treatment of the problem. 
First, we assume that the cues generated by the system are limited to a predefined set of \emph{atomic cues}. 
Second, as with the other scenarios, we assume that the subject being observed belongs to a closed set of subjects, whose examples of response to the generated cues were recorded in the preceding training phase. The subjects are free to respond with any \emph{response instances} (action instances) they find convenient. 
Finally, we assume that we can categorize the response instances into a closed set of \emph{atomic action response types} defined implicitly (see Section~\ref{sec:ResTypeRep}). 
At each step, the system is free to choose a single cue, from the set of atomic cues, expected to provide it with fast and accurate identification. 
To emphasize the problem's sequential nature we now use the same notations for the random variables as in Section~\ref{sec:GenModel} with subscripts $t$, 
representing the iteration index. Fig.~\ref{fig:IterativeProcess}(a) presents the generative model of the action instance creation process.

\subsection{Online Interactive Classification}
\label{sec:active}
We now define the interactive identification problem more formally.
Since a single subject is observed throughout the entire interactive session, our goal is to estimate $u$, which is the system's time independent state. Similarly to~\cite{DenzlerBrown}, we formulate our problem of interactive identification as an iterative process striving to correctly estimate $u$ using as few iterations as possible. Thus, the goal of each iteration is to reduce as much as possible the estimator's uncertainty, expressed as the distribution's deviation from the Dirac delta function.

\begin{figure}[tb]
\caption{Iterative CPI process. A cue is displayed; the person responds; the response updates the identity posterior; the next cue is chosen to maximise information gain.}\label{fig:IterativeProcess}
\end{figure}

At the beginning of iteration $t$, the probability distribution $p_{t-1}(u)$ represents the system's current belief in its state. 
The goal of the system is to select a cue $c_t$, such that the subject's response to it is expected to reduce the uncertainty in $u$. 
The subject's response comprises two factors -- the response type $a_t$ and the response instance $\mathbf{x}_t$.
The reduction in $u$'s uncertainty, due to an expected observation $\mathbf{x}_t$ obtained as a response to $c_t$, 
is defined as the \emph{conditional mutual information}~\cite{CoverThomas} between $u$ and $\mathbf{x}_t$, given $c_t$, and is denoted by $I(u;\mathbf{x}_t|c_t)$\footnote{Sometimes the same quantity is referred to as the \emph{information gain} ($IG$).}.
Thus, the optimal $c_t^*$ is selected by: 
\label{eq:IG-Max}
\begin{align*}
c_t^* &= \argmax_{c_t\in\mathcal{C}} I(u;\mathbf{x}_t|c_t),
\end{align*}

where (see Appendix~\ref{appendix:IG} for details),
\label{eq:IG-Max2}
\begin{align*}
I(u;\mathbf{x}_t|c_t) &= H(u|c_t) - H(u|\mathbf{x}_t,c_t) \\
                      &= \sum_{u\in\mathcal{U}}p(u)\int_{\mathcal{X}}p(\mathbf{x}_t|c_t,u) \log\frac{p(\mathbf{x}_t|c_t,u)}{p(\mathbf{x}_t|c_t)}d\mathbf{x}_t. 
\end{align*}

Using the generative model from Fig.~\ref{fig:IterativeProcess}(a), the joint likelihood of the response instance and type is factored into two components, each responsible for a different aspect of the subject's response. Thus, marginalizing over the response types, we obtain:
\begin{equation}
p(\mathbf{x}_t|c_t,u)=\sum_{a_t\in\mathcal{A}}p(\mathbf{x}_t,a_t|c_t,u)=\sum_{a_t\in\mathcal{A}}p(a_t|c_t,u)p(\mathbf{x}_t|a_t,u),
\end{equation}

where $p(a_t|c_t,u)$ models the likelihood of the subject's intention to respond in a certain way, while $p(\mathbf{x}_t|a_t,u)$ models the likelihood of the actual motion pattern being produced, once the intention is ``decided'' upon.

After the selected cue $c_t^*$ is generated and the subject's response $\mathbf{x}_t$ is observed, we use Bayes' rule to combine the prior with the likelihood to get the \emph{a~posteriori} distribution over the identities, which will serve as a prior for iteration $t+1$: 
$p_t(u)=p_t(u|c_t^*,\mathbf{x}_t)\propto p_{t-1}(u)p(\mathbf{x}_t|c_t^*,u)$. 
Fig.~\ref{fig:IterativeProcess}(b) visualizes the iterative process. Once the iteration count $t$ exceeds the maximally allowed number of iterations $T$, or the a~posteriori probability $p_t(u)$ of one of the identities exceeds a predefined threshold $th$, the process terminates, returning the maximum \emph{a posteriori} (MAP) estimate, $u^*=\argmax_{u\in\mathcal{U}}p_t(u)$. Algorithm~\ref{alg:CPI} summarizes the described process.

\begin{algorithm}
\caption{Cued Person Identification (CPI)}
\label{alg:CPI}
\begin{algorithmic}
\STATE{\textbf{Input:} $T$: max steps; $th$: threshold; $p(a|c,u)$, $p(\mathbf{x}|a,u)$: likelihoods; $p_0(u)$: prior; sets $\mathcal{C}$, $\mathcal{A}$, $\mathcal{U}$.}
\STATE{\textbf{Output:} $u^*$}
\WHILE{$t \leq T$ \textbf{and} $\max_{u}p_t(u)<th$}
  \STATE Select cue $c_t^*$ via Eq.~\ref{eq:IG-Max}.
  \STATE Obtain $\mathbf{x}_t$ in response to $c_t^*$.
  \STATE $p_{t}(u)\leftarrow p_{t-1}(u)\sum_{a_t}p(a_t|c_t^*,u)p(\mathbf{x}_t|a_t,u)$.
  \STATE $t \leftarrow t+1$.
\ENDWHILE
\STATE $u^*=\argmax_{u}p_t(u)$.
\end{algorithmic}
\end{algorithm}


In~\cite{DenzlerTR}, the authors prove that such a process converges to the correct solution by observing that it actually forms a Markov chain and assuming that it has only one stationary distribution.

\subsection{Training}
\label{sec:training}
The purpose of the training phase is to estimate the action type and instance likelihood functions, required for the computation of $p(\mathbf{x}|c,u)$.
This requires recording several examples of each subject $u\in\mathcal{U}$ reacting to each of the cues in $\mathcal{C}$. 
The reaction is composed of the response type $a$ and the response instance $\mathbf{x}$. 
Similar to the Login and the Smart home scenarios, we use 1-NN KDE (Eq.~\ref{eq:kde}) to estimate $p(\mathbf{x}|a,u)$.
We estimate $p(a|c,u)$ by computing the relative frequency of subject $u$ responding with $a$ to cue $c$,
\begin{equation}
\label{eq:rcMap}
p(a|c,u) = \frac{\#(\text{$u$'s responses with $a$ to $c$})}{\#(\text{total $u$'s responses to $c$})}.
\end{equation}

\subsection{Evaluation on Toy Data}
To demonstrate the intuition behind the CPI algorithm we demonstrate its performance on randomly generated toy data.
Consider a set of $N$ atomic response types, $\mathcal{A}_c=\mathcal{C}$, defined by the set of atomic cues, $\mathcal{C}=\left\{1,...,N\right\}$, and a set of $2^N$ user identities, $\mathcal{U}=\{0,1\}^{N}$.
We define a univariate probability distribution of the action instance $x\in\mathbb{R}$ given the response label $a\in\mathcal{A}$ and the person identity $u\in\mathcal{U}$ as 
$p(x|a,u)\sim\mathcal{N}(u(a),\sigma^2(a))$, where $u(a)$ is the $a$'th bit of the user's identifier $u$ and $\sigma(a)=\sqrt{s\cdot a}$ is the standard deviation of the 
$a$'th instances depending on a positive scalar $s$. 
Note that this particular choice of distributions, where as $a$ increases so does $\sigma(a)$, makes it harder to determine the true value of the user identifier's least significant bits.

\begin{figure*}[tb]
\begin{center}
\begin{tabular}{ccc}
\includegraphics[width=5.2cm]{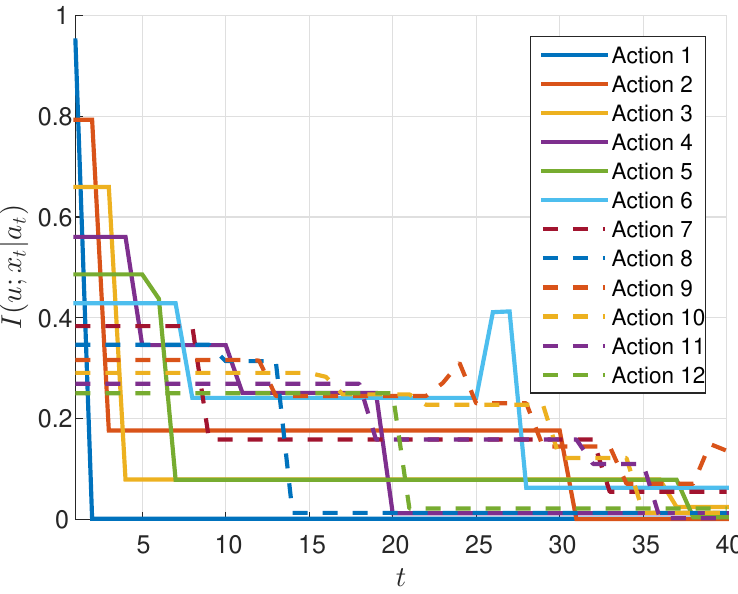} &
\includegraphics[width=5.2cm]{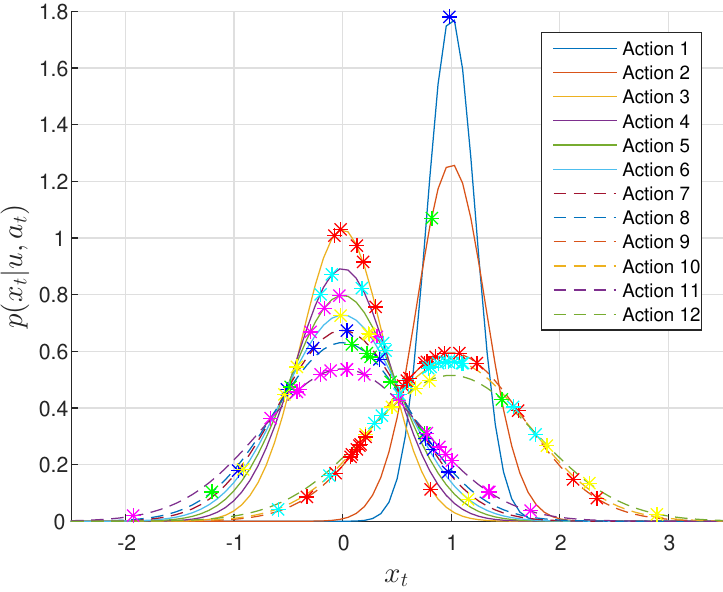} &
\includegraphics[width=5.2cm]{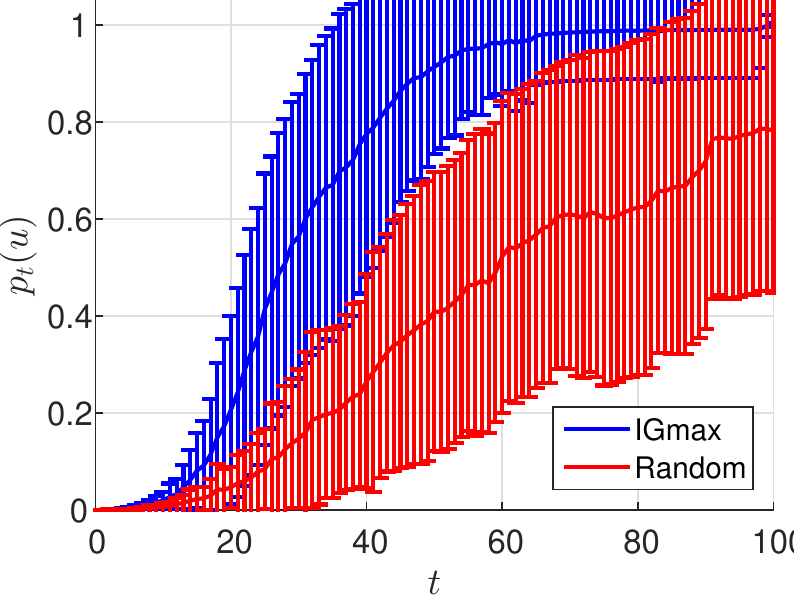}  \\
(a) & (b) & (c) 
\end{tabular}
\caption{\textbf{Toy Data.}
(a) Information gain expected from different response types. Note the immediate drop in the information gain of response $a$ once its instance is observed.
(b) The color-coded ``*'' correspond to the generated action instances (one instance per iteration). See text for details.
(c) The performance of the CPI algorithm using the IG-Max and the Random selection criteria evaluated in terms of the true class probability.
}
\label{fig:IG-MaxVsRandom}
\end{center}
\end{figure*}

Note that generating a cue $c'$ guarantees that we will receive back information regarding the $c'$'th bit of the user's identifier.
The parameter values in this experiment are $N=12$ and $s=0.05$. 
The inputs to the algorithm are $T=100$, $th=1$, $p_0(u)=\frac{1}{2^N}, \forall u\in\mathcal{U}$, and $p(a|c,u)$, $p(x|a,u)$, $\mathcal{A}$, $\mathcal{U}$ and $\mathcal{C}$ are as defined above. 
To simulate the users' responses we generated a response instance $x_t$ using the distribution $p(x_t|a_t,u')$, where $a_t$ is the selected response in step $t$ and $u'$ is the user we are trying to identify. 
Fig.~\ref{fig:IG-MaxVsRandom}(a) shows the information gain expected from cues in $\mathcal{C}$ during the first 40 execution steps, tested on a randomly chosen user identifier 
$u'=110000001101$. Note, for example, that at $t=2$, ``Action 2'' (solid red line) was observed for the first time (by generating ``Cue 2'') since at $t=1$ the information gain of ``Cue 1'' (corresponding to ``Action 1'') was still higher. 
The observation resulted in a sharp drop in the information gain of ``Cue 2'', which was selected again only at $t=30$. To demonstrate the role of parameter $\sigma(a)$, we note that during 100 steps of the algorithm, ``Cue 2'' was selected only twice, while ``Cue 11'' was selected 17 times.  
Fig.~\ref{fig:IG-MaxVsRandom}(b) shows the actually generated instances $\{x_t\}_{t=1}^T$, color-coded according to their corresponding response labels $\{a_t\}_{t=1}^T$, overlaid on top of their original distributions. Note that the number of points generated from the distribution of response $a$ is proportional to the standard deviation 
$\sigma(a)$. 

In our experiments we averaged the results over 20 randomly chosen user identities tested five times each (overall 100 test runs). 
We compared the IG-Max action selection strategy with the random strategy, selecting the cues at random.
Fig.~\ref{fig:IG-MaxVsRandom}(c) shows the probability of the true user identity, $p_t(u')$, as a function of $t$ for both strategies. 
Note that, on average, when $t>50$, it takes Random at least twice the number of iterations to reach the same level of confidence.

\subsection{Relation to Human Information Processing}
Based on our generative model, the likelihood of the user's response to a specific cue can be decomposed into the likelihood of the user's intent (psychological) and the likelihood of the actual motion pattern being produced to fulfill the intent (physiological).
The choice of the generative model was partially motivated by the Human Information Processing (HIP) paradigm, well known in psychology, and more specifically, in the human--computer interaction (HCI) domain~\cite{HCISearsJacko}. According to this approach, the user's brain processes the displayed stimulus in three consecutive stages.
In the first, the perception stage, the stimulus is identified. In the second, the cognition stage, the response is selected. 
Finally, in the third, the execution stage, the response is executed. All three stages are quite complicated to model; however, by following the black box approach it is possible to represent each stage by modeling the probabilities of obtaining certain outputs given certain inputs. The first two stages are represented by 
$p(a|c,u)$, modeling the individual's perceptual and cognitive capabilities. The third stage is represented by $p(\mathbf{x}|a,u)$, modeling the individual's motoric capabilities. The underlying assumption is that, given enough labeled data, these empirically learned distributions will eventually describe the true ones accurately enough. Fig.~\ref{fig:IterativeProcess}(c) illustrates the connection between our generative model and the HIP paradigm.  

\section{Interactive Scenario Experimental Setup}
\label{sec:system}
\begin{figure*}[tb]
\begin{center}
\begin{tabular}{ccccc}
\includegraphics[width=3.5cm]{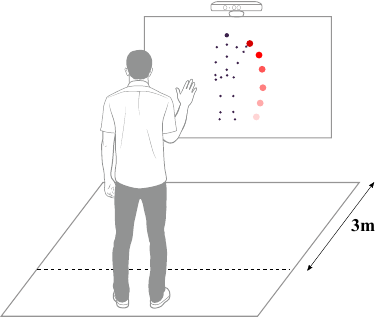} &
\includegraphics[width=2.5cm]{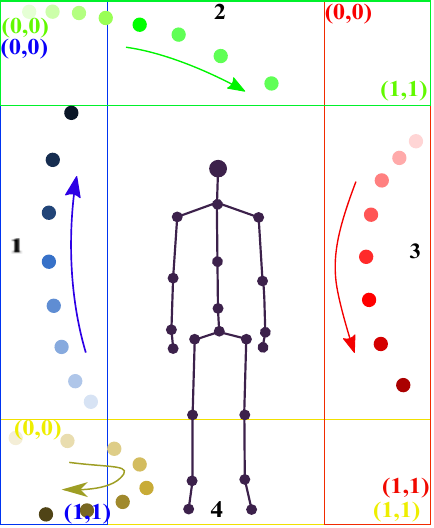} &
\includegraphics[width=3.5cm]{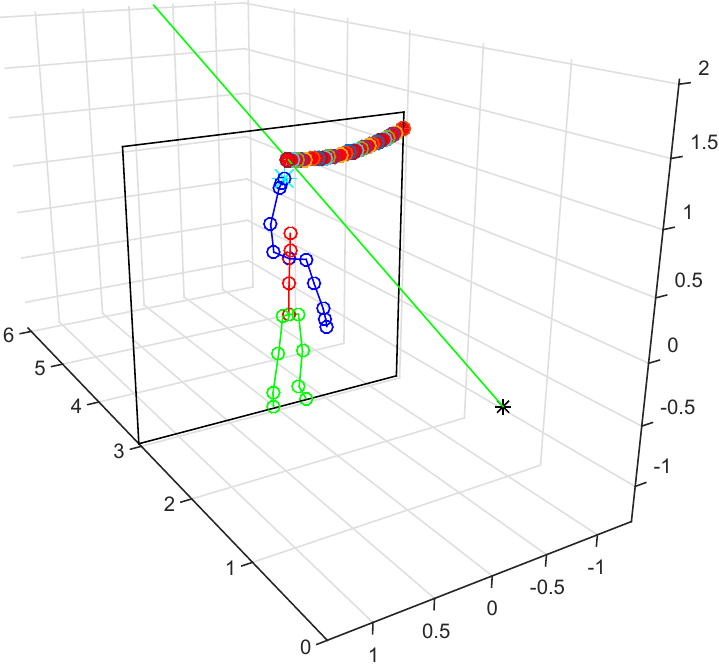}  &
\includegraphics[width=3.5cm]{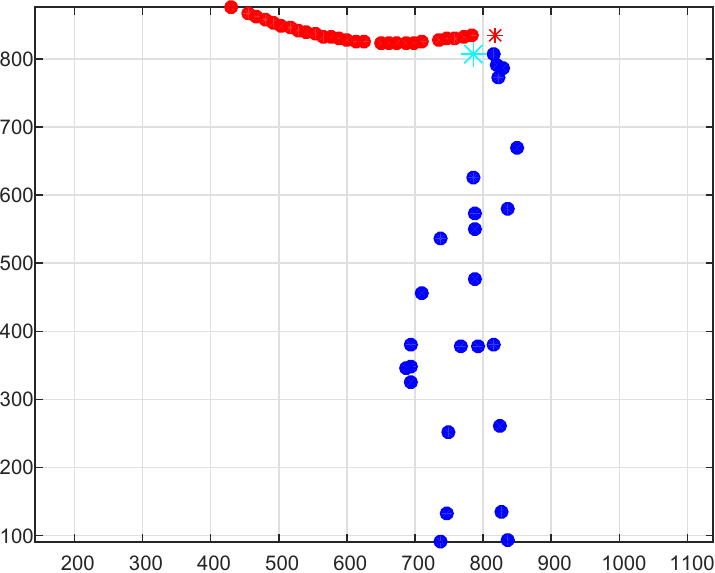} & 
\includegraphics[width=3.5cm]{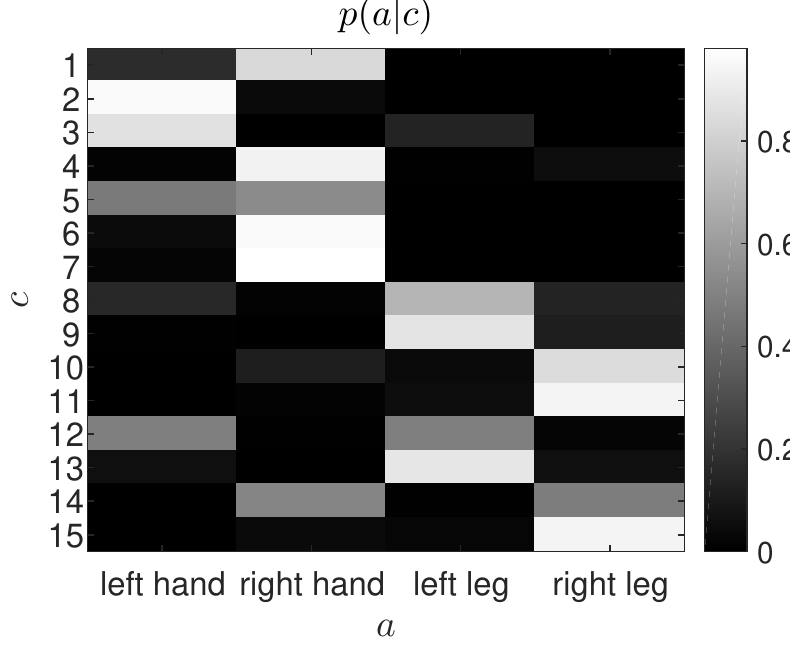}\\
(a) & (b) & (c) & (d) & (e) \\
\end{tabular}
\caption{CuedId dataset collection setup.
(a) During the game the subject sees his or her skeleton live on the screen together with the rendered cue, projected from 3D. 
(b) Four regions for cue generation and example cues.
(c) The subject and the cue trajectory in 3D. The black frame represents the subject's frontal plane, and the green ray connects the camera center with the interception point.
(d) The screen reflecting the interception moment as seen by the player. Cyan asterisk highlights the intercepting joint. 
(e) Average probability to observe a response type $a$ to cue $c$.
}
\label{fig:CuedId}
\end{center}
\end{figure*}

To evaluate the described methods on real data, we developed an interactive framework based on a Kinect game that can be used for data collection and testing the approach in real-time.
The hardware setup includes a Kinect 2 sensor mounted on top of a large screen, both of which are connected to a machine running the game.
In the course of a single game, a player is shown a sequence of stimuli (cues) either in a predefined or random order. 
The player's response to each cue is recorded and stored in the database. 
Skeletal joints are tracked in 3D, projected onto the screen and rendered as a set of markers in 2D coordinates in real-time. 
The cue, represented as a trajectory of a 3D point in time, is defined by its initial position, velocity and acceleration, and a gravity field governing its dynamics. Cues are visualized as 2D flying asteroids projected onto the screen along with the player's skeleton. 
The player is supposed to intercept the asteroids with any limb he or she finds convenient while moving freely.
They, however, are not allowed to get closer than 3 meters from the screen. 
The interception is defined as a collision of any joint marker with the asteroid bitmap in the 2D plane of the screen\footnote{We noticed that visualization in 2D is more intuitive for the user than in 3D, since it is no longer necessary to resolve the joints' depth with respect to that of the asteroid.}. 
The response to a given cue is defined as the player's 3D skeletal motion pattern recorded from the moment the asteroid appears on the screen until it is intercepted. 
In case of a miss (we allow the asteroids to get outside the screen boundaries), the cue is regenerated.
Fig.~\ref{fig:CuedId}(a) illustrates the setup and the game. Note that a joint-cue collision in 2D is equivalent to a joint in 3D intercepting a ray, connecting the camera center with the asteroid's current 3D position (see Fig.~\ref{fig:CuedId}(c)).
 
We chose the set of cues so as to stimulate subjects to respond in various ways while the cue parameters were empirically selected in a trial and error fashion. 
To equally stimulate the proprioceptive spaces of different individuals, regardless of their physical dimensions or positions with respect to the screen, the cues' parameters are defined in the subject's coordinate systems. We define four rectangular regions adjacent to the bounding box around the player's silhouette. Each region has its own ID and a coordinate system for defining the cue's initial parameters (see Fig.~\ref{fig:CuedId}(b)).   
Each cue is defined by a set of 10 parameters. The set of parameters is $\left(r,\mathbf{p}_0,\mathbf{v}_0,\mathbf{a}_0\right)$, where $r\in\left\{1,\ldots,4\right\}$ is the region ID and $\mathbf{p}_0,\mathbf{v}_0,\mathbf{a}_0\in\mathbb{R}^3$ are the trajectory initial point's position, velocity and acceleration, respectively. 
For our experiments we constrained the cues' trajectories to lie in the subject's frontal (coronal) plane; therefore, the $z$ component of $\mathbf{p}_0$ is determined by the subject's center of mass and the $z$ components of $\mathbf{v}_0$ and $\mathbf{a}_0$ are set to 0.   
For example, a red cue in Figs.~\ref{fig:CuedId}(c-d) is defined by the following parameters: $\left(2,\left[0,0,0\right]^T,\left[50,20,0\right]^T,\left[0,-120,0\right]^T\right)$.

\subsection{CuedId Dataset}
\label{sec:CuedIdDataset}
We recorded 22 subjects playing at least nine games each, where each game comprises a sequence of 15 cues displayed in either predefined or randomly shuffled order.
This results in 4,476 recordings overall. The recordings were made over the course of three months using the system described in Section~\ref{sec:system}.
The subjects include both genders (15 males and 7 females) ranging from 19 to 60 years of age.
We instructed the subjects to return to the initial position (upright with hands down, facing the screen) after intercepting an asteroid.
The dataset consists of cue 3D trajectories -- sequences of cue positions in 3D -- synchronized with the sequence of the player's 3D poses. Additionally, the dataset contains
the corresponding 2D projections of the cue trajectories and the poses as were observed by the player on the screen.

\subsection{Action Type Representation}
\label{sec:ResTypeRep}
According to our generative model, the latent variable $a$ takes values from a predefined set of atomic response types 
$\mathcal{A}$. Since we instruct the users to perform the most convenient gesture for intercepting the cue, we must define $\mathcal{A}$ implicitly rather than explicitly.
The most straightforward definition of the response type set is $\mathcal{A}_c=\mathcal{C}$. Thus, the response $a$ is defined simply as ``the response to cue $c$''.
This definition implies a one-to-one correspondence between the response types and the cues, turning $a$ degenerate. Consequently, to learn $p(\mathbf{x}|c,u)$, the subject $u$ has to be sampled using each of the cues $c\in\mathcal{C}$, which is not practical, especially for very large $\mathcal{C}$. In real-life interactive applications, the number of response types is usually much smaller than the number of possible cues ($|\mathcal{A}|<<|\mathcal{C}|$), and therefore, designating a set of plausible responses makes the learning process more manageable. 

Another drawback of the straightforward definition is that it does not take advantage of the actual user--cue interaction revealing the subject's original intent. 
For a given cue, due to the physical constraints of the human body, a response of an average person is most likely to be one of a small portion of all possible response types. 
Thus, for cue $c'$ and two users $u_1$ and $u_2$, the probability vectors $p(a|c',u_1)$ and $p(a|c',u_2)$ are likely to have identical support.
Nonetheless, different subjects have their own preferences and capabilities, and therefore, for a given cue, the probabilities of performing each of the response types from the support varies.
For example, an object passing near one's torso will most likely be intercepted by one of the upper limbs. 
However, whether this will be the right or left limb depends, among other things, on whether the subject is left or right-handed.
Therefore, it is reasonable to define the response types as a set of potential limbs used for interaction: $\mathcal{A}_l=\{$left arm, right arm, left leg, right leg$\}$.
In our experiments, we report the results for both $\mathcal{A}_c$ and $\mathcal{A}_l$.

Since the number of possible responses to a specific cue is limited by the physical capabilities of the human body, the support of $p(a|c)$ is likely to be sparse. This statement is validated in our experiments. Fig.~\ref{fig:CuedId}(e) shows the values of
$p(a|c)$ for each potentially intercepting limb ($a\in\mathcal{A}_l$). Note that $c\in\{1,\ldots,7\}$ is more likely to be intercepted with one of the hands, while
$c\in\{8,\ldots,15\}$ is more likely to be intercepted with one of the legs. This is so because in the first group of cues the asteroid is moving towards the upper part of the body, \eg,
green and blue regions in Fig.~\ref{fig:CuedId}(b), while in the second group the asteroid is moving towards the lower part, \eg, red and yellow regions in Fig.~\ref{fig:CuedId}(b).
  

\section{Experimental Results}
\label{sec:res}
We divide our experiments into two groups -- one containing the Login and the Smart home scenarios, and the second one, containing the Interactive scenario. In the former, we evaluate our approach for identifying a person based on contextual motion on public datasets and benchmark it against publicly reported baselines. In the latter,    
we focus on evaluating the merits of the Interactive approach with respect to the baseline Random cue selection strategy.

\subsection{Login and Smart Home Scenarios}
In our experiments we use seven datasets containing contextual skeletal motion patterns performed by different individuals. The datasets include five publicly available ones in the action recognition domain, \ie, the NTU RGB+D~\cite{shahroudy2016ntu}, MSRC-12~\cite{MSRC12}, UCFKinect~\cite{UCFKinect}, MSR-Action3D~\cite{3DBag} and UTKinect~\cite{UTKinect} datasets, one dataset for person authentication, \ie, the BodyLogin~\cite{BodyLogin} dataset, and our own privately collected CuedId dataset described in Section~\ref{sec:CuedIdDataset}.

\subsubsection{Evaluation Protocol}
Most of the existing human motion 3D skeletal datasets in the action recognition domain contain labels of subjects performing the action. 
This is done to prevent action recognition algorithms from overfitting on actions of specific subjects during training. 
We use these labels for our task. Table~\ref{tab:datasets} presents a summary of all the contextual motion datasets used in our experiments. 
For each dataset, the number of actions (contextual motion types), users, available user-action instances and the total number of available instances 
are specified.


We differentiate between the Login and the Smart home scenarios by assuming that the action class label is given or inferred, respectively. 
We considering two cases:
\begin{enumerate}
\item \textbf{Login.} Assume that the action label is given by the dataset ground truth.
\item \textbf{Smart home.} Assume that the action labels are unknown, but may be inferred using any given action
classifiers, \eg~\cite{Vemulapalli}, as discussed in Section~\ref{sec:ActReco}. 
\end{enumerate}
In all our experiments we report recognition results averaged over 10 random selections of action instances into training and test sets.

To evaluate the benefits of action-specific metric learning we use two metric types:
\begin{enumerate}
\item \textbf{L2.} No action-specific metric learning is performed, \ie, $\mathbf{M}_a = \mathbf{I}, \forall a\in\mathcal{A}$.
\item \textbf{Mahalanobis} A Mahalanobis distance metric $\mathbf{M}_a$ is learned using the dimensionality reduction as described
in Section~\ref{sec:distMetric}. We set $D_p = 30$ and $D_l = 15$ in all our experiments, except for the BodyLogin dataset, 
where we set $D_p = 40$ and $D_l = 39$.
\end{enumerate}


\begin{table*}[ht]
\small
\begin{center}
\begin{tabular}{|l||c|c|c|c|c|}
\hline
\textbf{Dataset} & \textbf{Users} & \textbf{Contexts} & \textbf{Inst. Per} & \textbf{Inst.} & \textbf{Contextual Motion Types} \\
\textbf{Name} & \textbf{Num.} & \textbf{Num.} & \textbf{User-Cont} & \textbf{Tot.} &  \\ \hline
\textbf{NTU RGB+D}~\cite{shahroudy2016ntu} & 40 & 60 & 3-96 & 56880 & http://rose1.ntu.edu.sg/Datasets/actionRecognition.asp \\ \hline

\textbf{MSRC-12}~\cite{MSRC12} & 29 & 12 & 8-10 & 6015 & 
\begin{tabular}{@{}c@{}} 
\footnotesize 1.Lift outstretched arms; 2.Duck; 3.Push right; 4.Goggles; 5.Wind it up; 6.Shoot; \\
\footnotesize 7.Bow; 8.Throw; 9.Had enough; 10.Change weapon; 11.Beat both; 12.Kick \end{tabular} \\ \hline

\textbf{UCFKinect}~\cite{UCFKinect} & 16 & 16 & 5 & 1280 & 
\begin{tabular}{@{}c@{}}
\footnotesize 1.Balance; 2.Climb ladder; 3.Climb up; 4.Duck; 5.Hop; 6.Kick; 7.Leap; \\
\footnotesize 8.Punch; 9.Run; 10.Step back; 11. Step front; 12. Step left; \\
\footnotesize 13. Step right; 14. Twist left; 15. Twist right; 16. Vault
\end{tabular} \\ \hline

\textbf{MSR-Action3D}~\cite{3DBag} & 9 & 16 & 2-3 & 427 & 
\begin{tabular}{@{}c@{}}
\footnotesize 1.High arm wave; 2.Hor. arm wave; 3.Hammer; 4.Forward punch; \\
\footnotesize 5.High throw; 6.Draw X; 7.Draw tick; 8.Draw circle; 9.Hand clap; \\
\footnotesize 10.Two hand wave; 11.Side boxing; 12.Forward kick; 13.Jogging; \\
\footnotesize 14.Tennis swing; 15.Tennis serve; 16.Golf swing
\end{tabular}
\\ \hline
\textbf{UTKinect}~\cite{UTKinect} & 9 & 10 & 2 & 180 & 
\begin{tabular}{@{}c@{}}
\footnotesize 1.Walk; 2.Sit down; 3.Stand up; 4.Pick up; 5.Carry; 6.Throw; \\
\footnotesize 7.Push; 8.Pull; 9.Wave hands; 10.Clap hands
\end{tabular}
\\ \hline 
\textbf{BodyLogin}~\cite{BodyLogin} & 40 & 1 & 20 & 800 & 
\footnotesize 1. The ``S'' action \\ \hline
\textbf{CuedId} (Ours) &  22 & 15 & 10--15 & 4476 &  
\footnotesize Limb motion patterns of intercepting virtual objects in space (see Fig.~\ref{fig:CuedId}(b)).
\\ \hline

\end{tabular}
\end{center}
\caption{\label{tab:datasets} Summary of contextual motion datasets used in our experiments.}
\label{tab:datasets}
\end{table*}



\subsubsection{Our Approach Evaluation}
In all our experiments we classify the user at test time using a single randomly selected action instance ($N=1$). The number of training instances vary with dataset. For the MSRC-12 and the UCFKinect datasets, we randomly choose four action instances belonging to each possible user-action combination into the training set. 
We report results using both the L2 distance metrics and the Mahalanobis distance metrics, where applicable. For the MSR-Action3D and the UTKinect datasets only one action instance is used for training due to the limited dataset size.  

In all of our experiments in this section, unless stated otherwise, we use the DTW+FTP to perform temporal normalization (described in Section~\ref{sec:TempNorm}). For more details and ablation studies please refer to~\cite{Kviatkovsky_2015_CVPR_Workshops}. For the interactive scenario, evaluated in Section~\ref{sec:res_cued_id}, we use cubic spline interpolation. 

Table~\ref{tbl:ReIdTPR_tst} summarizes the True Positive Rate (TPR) for each dataset.
Note also that, where applicable, as expected, the Mahalanobis
distance metric outperforms the L2. We also see that the recognition accuracy degrades gracefully with the uncertainty introduced by action recognition (in the transition between the Login and Smart home scenarios).
Trying to fit the CuedId dataset into the Login and Smart home scenarios, we speculate that using $\mathcal{A}_c$ as the response action types is similar to the Login scenario, while using $\mathcal{A}_l$
turns the experience into more resembling the Smart home scenario, since the response action category has to be inferred based on the action instance. 
It can be clearly seen that the recognition accuracy increases with the semantic complexity of the performed action. For example, for the MSRC-12 and the BodyLogin datasets, containing semantically rich contextual motions, the recognition accuracy is 97\% and 95.7\%, respectively. In contrast, for the UTKinect and CuedId datasets, containing much more primitive contextual motions, the recognition accuracy is significantly lower -- 58\% and 65.2\%, respectively.
Thus, we see that semantic constraints enforced on the range of motion, do improve the reliability of the identification modality. However, this comes with a price of a less natural experience. In section~\ref{sec:res_cued_id} we show that it is possible to improve the accuracy using an ensemble of weaker modalities, with no degradation in the user experience.


\begin{table*}[ht]
\small
\begin{center}
\begin{tabular}{|c|l|c|c|c|c|c|c|c|c|}
\hline
 & \multicolumn{2}{|c|}{\textbf{MSRC12}~\cite{MSRC12}} & \multicolumn{2}{|c|}{\textbf{UCFKinect}~\cite{UCFKinect}} & \multicolumn{1}{|c|}{\textbf{MSRAction3D}~\cite{3DBag}} & \multicolumn{1}{|c|}{\textbf{UTKinect}~\cite{UTKinect}} & \multicolumn{1}{|c|}{\textbf{BodyLogin}~\cite{BodyLogin}} & \multicolumn{1}{|c|}{\textbf{CuedId}} \\
\hline
 \textbf{Scenario} &  $\mathbf{L}_2$ & \textbf{Mahal.} &  $\mathbf{L}_2$ & \textbf{Mahal.} & $\mathbf{L}_2$ & $\mathbf{L}_2$ & \textbf{Mahal.} & $\mathbf{L}_2$ \\ 
\hline
\hline
\textbf{Login} & 97 & \textbf{98} & 92 & \textbf{96} & 93 & 58 & 95.7 & 65.2 ($\mathcal{A}_c$)\\ \hline
\textbf{Smart Home} & 93 & \textbf{95} & 91 & \textbf{95} & 90 & 57 & - & 53.8 ($\mathcal{A}_l$)\\ \hline
\end{tabular}
\end{center}
\caption{Average TPRs for different scenarios evaluated on various contextual motion datasets. The highest result for each dataset and scenario combination is highlighted.}
\label{tbl:ReIdTPR_tst}
\end{table*}

\subsubsection{Comparing to Classic State-Of-The-Art Methods}
In this section we compare our method to the state-of-the-art in person identification from motion patterns, using classic machine learning approaches. 
In~\cite{ActionStylesRF}, the authors propose a context-agnostic approach for action style recognition, where the set of actions used in training 
is not necessarily identical to that used in testing. As part of their experimental evaluation, the authors compare the performance of many popular classifiers on the same datasets we used in~\cite{Kviatkovsky_2015_CVPR_Workshops} and we use here. The authors use the same evaluation protocol making our results directly comparable. Table~\ref{tbl:Comparison1} presents the average TPRs for our approach compared to the Hough Forest (HF-R)~\cite{gall2011hough, ActionStylesRF}, the 3D Joints' Covariance Descriptor (Cov3DJ)~\cite{HusseinCovIJCAI, BodyLogin}, and the Discrete Time Warping (DTW)~\cite{Lai, Wu} approaches. HF-R~\cite{ActionStylesRF} uses a na\"ive Bayes assumption on the pose sequence, rendering pose features independent given the user and action labels. This results in a bag-of-poses-like~\cite{UCFKinect} approach, discarding the temporal information contained in the motion pattern. Interestingly, it outperforms all methods which do take temporal information into account. However, we do not think that such an approach may be scalable to a large number of subjects, as the discriminative power of anthropometric features is expected to decrease with an increased number of identities. Among the remaining methods, DTW~\cite{Wu} has the highest score and is probably the method of choice given a moderate size enrollment set. Note that in test time, a DTW distance is computed between the query sequence and each one of the enrolled sequences, significantly increasing the identification latency.
Our method outperforms Cov3DJ~\cite{BodyLogin} by a large margin, making it the most appealing for real time applications, such as, for example, the Interactive scenario.

\begin{table*}[ht]
\small
\begin{center}
\begin{tabular}{|c|l|c|c|c|c|c|}
\hline
\textbf{Approach} & \textbf{MSRC12}~\cite{MSRC12} & \textbf{UCFKinect}~\cite{UCFKinect} & \textbf{MSRAction3D}~\cite{3DBag} & \textbf{UTKinect}~\cite{UTKinect} \\
\hline
\hline
\textbf{Cov3DJ}~\cite{BodyLogin} & 83.74 & 74.84 & 78.83 & 29.22 \\ \hline
\textbf{DTW}~\cite{Wu} & 97.54 & 96.13 & 97.55 & 64.00 \\ \hline
\textbf{Smart Home, DTW+FTP+Mahal.} & 95.00 & 95.00 & 90.00 & 57.00 \\ \hline
\textbf{HF-R}~\cite{ActionStylesRF} & \textbf{99.44} & \textbf{99.65} & \textbf{99.78} & \textbf{73.00} \\ \hline
\end{tabular}
\end{center}
\caption{Comparison to several baselines reported in the literature.}
\label{tbl:Comparison1}
\end{table*}

Additionally to pose-based approaches, the authors of~\cite{ActionStylesRF} evaluate several popular classifiers with the exact same action representation used in~\cite{Kviatkovsky_2015_CVPR_Workshops} and also used here -- the joint position (JP) representation. This allows us to make a direct comparison of our classification scheme to possible, more complex, alternatives. Table~\ref{tbl:Comparison2} presents a comparison of our method, \ie, DTW+FTP+Mahal., to four classifiers, \ie, K-Nearest Neighbors (kNN), Support Vector Machines (SVM) with linear (L) and radial basis function (R) kernels, and Random Forest (RF), on four public datasets (for details, please refer to~\cite{ActionStylesRF}). Our approach outperforms kNN and both SVMs for all datasets. The RF classifier slightly outperforms our approach on the UCFKinect and the MSRAction3D datasets, but on average it performs on par with our approach. We conclude that, although our classification scheme is simpler than the alternative classifiers, the proposed generative model compensates for it, achieving high accuracy combined with low training complexity.   

\begin{table*}[ht]
\small
\begin{center}
\begin{tabular}{|c|l|c|c|c|c|c|}
\hline
\textbf{Approach} & \textbf{MSRC12}~\cite{MSRC12} & \textbf{UCFKinect}~\cite{UCFKinect} & \textbf{MSRAction3D}~\cite{3DBag} & \textbf{UTKinect}~\cite{UTKinect} \\
\hline
\hline
\textbf{kNN} & 84.11 & 84.06 & 68.00 & 46.89 \\ \hline
\textbf{SVM-L} & 91.99 & 90.35 & 87.69 & 41.89 \\ \hline
\textbf{SVM-R} & 93.97 & 93.16 & 87.12 & 46.11 \\ \hline
\textbf{RF} & 94.09 & \textbf{98.09} & \textbf{90.68} & 46.67 \\ \hline
\textbf{Smart Home, DTW+FTP+Mahal.} & \textbf{95.00} & 95.00 & 90.00 & \textbf{57.00} \\ \hline
\end{tabular}
\end{center}
\caption{Average TPRs of our approach compared to the ``holistic methods'' reported in the follow-up work~\cite{ActionStylesRF}.}
\label{tbl:Comparison2}
\end{table*}

\subsubsection{Comparing to DL-based State-Of-The-Art Methods}
Although deep learning has revolutionized skeleton-based action recognition, to the best of our knowledge,~\cite{Wang&Wang} is the only work which followed the trend in the domain of person identification from everyday action skeleton motion sequences. This might be due to the lack of appropriate datasets until recently.
In~\cite{Wang&Wang} the authors present several deep Recurrent Neural Net (RNN) architectures trained in a multi-task setting, jointly supervised by the action and subject labels. The evaluation is done on the NTU RGB+D~\cite{liu2019ntu} datset.

Independently Recurrent Neural Net (IndRNN)~\cite{li2018independently} is a recently introduced variation of RNN, which is much easier and faster to train. Besides the practical aspects, it has significantly outperformed RNN and LSTM on the skeleton-based action recognition. 
To assess the performance of our approach in the deep learning era, we trained and evaluated an IndRNN model on two of the largest available datasets, \ie, MSRC-12~\cite{MSRC12} and NTU RGB+D~\cite{shahroudy2016ntu}, and compared to the performance our approach. For both datasets, we used the 6-layer simple IndRNN architecture described in~\cite{li2018independently}. 
Table~\ref{tbl:DL_Comparison2} shows the average TPR of IndRNN on the MSRC-12 dataset, compared to several variations of our approach. Albeit the increased complexity of the classification scheme, IndRNN's average TPR of 92.76\% is similar to that of our approach. We note that the comparison might not be completely fair as the IndRNN model was trained with the subject ID as the only supervision channel, as opposed to our approach, which also utilized the action label. 

As of today, NTU RGB+D~\cite{shahroudy2016ntu} is the largest and the most varied dataset of everyday actions with a total of 56,880 skeletal sequences of 60 action types performed by 40 individuals, captured form three different angles. We evaluated the IndRNN model on the NTU RGB+D dataset following the cross-view (CV) protocol, where the frontal views are used for training and the two half-side views are used for testing. We also evaluated our approach in the same setting. 
Table~\ref{tbl:DL_Comparison1} presents the average TPR of IndRNN, of our approach, and the results reported in~\cite{Wang&Wang} including several baseline methods.   
Again, we note that the IndRNN model was trained with the subject ID as the only supervision channel, as opposed to the best Multi-task RNN architecture (Late split, FC)~\cite{Wang&Wang}, jointly trained with action and subject label. Nevertheless, IndRNN achieves a TPR of 67.53\% outperforming the best reported model in~\cite{Wang&Wang}, which achieves a TPR of 65.2\%. For comparison, the TPR of a Multi-task RNN (Separate Net), trained with a subject ID supervision only is 47.6\%. The TPR of our approach on the NTU RGB+D dataset is 50.92\% and 41.37\% for the Login and the Smart home scenarios, respectively. The Smart home (Mahal.) results in a comparable performance to that of the 3-layer LSTM, outperforming two non-deep approaches, \ie, Skeleton Features~\cite{Barbosa:reid12} and Anthropometric Features~\cite{pala2015multimodal}. The Login (Mahal) outperforms two deep learning based approaches, \ie, the Multi-task RNN (Separate Net) and the 3-layer LSTM.

\begin{table}[ht]
\small
\begin{center}
\begin{tabular}{|c|l|c|}
\hline
\textbf{Approach} & \textbf{Accuracy} \\
\hline
\hline
$\mathbf{L}_2$ & 82.0  \\ \hline
\textbf{Mahal.} & 92.0  \\ \hline
\textbf{IndRNN} & 92.76  \\ \hline
\textbf{DTW+FTP+L}$_2$ & 93.0  \\ \hline
\textbf{DTW+FTP+Mahal.} & \textbf{95.0}  \\ \hline
\end{tabular}
\end{center}
\caption{Comparison of our approach to the IndRNN based approach on the MSRC12~\cite{MSRC12} dataset, assuming the Smart Home scenario.}
\label{tbl:DL_Comparison2}
\end{table}

\begin{table}[ht]
\small
\begin{center}
\begin{tabular}{|c|l|c|}
\hline
\textbf{Approach} & \textbf{Accuracy} \\
\hline
\hline
\textbf{Skeleton Features}~\cite{Barbosa:reid12} & 21.0 \\ \hline
\textbf{Anthropometric Measures}~\cite{pala2015multimodal} & 28.6 \\ \hline
\textbf{Smart Home, Mahal.} &  41.37 \\ \hline
\textbf{3 layer LSTM}~\cite{Wang&Wang} & 43.5  \\ \hline
\textbf{Multi-task RNN (Separate Net)}~\cite{Wang&Wang} & 47.6  \\ \hline
\textbf{Login, Mahal.} &  50.92 \\ \hline
\textbf{Multi-task RNN (Late Split FC)}~\cite{Wang&Wang} &  65.2  \\ \hline
\textbf{IndRNN} &  \textbf{67.53}  \\ \hline
\end{tabular}
\end{center}
\caption{Comparison to other approaches on the NTU RGB+D~\cite{shahroudy2016ntu} dataset. For action classification in the Smart home scenario, we have used an independently trained IndRNN, obtaining a TPR of 73.4\%, resulting in a significant accuracy reduction w.r.t. the Login scenario.}
\label{tbl:DL_Comparison1}
\end{table}

\subsubsection{Homogeneous vs. Heterogeneous Sets}
\label{sec:HomoVsHetero}
\begin{figure}[ht]
\begin{center}
\begin{tabular}{cc}
\includegraphics[width=4cm]{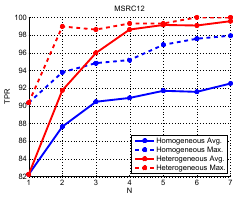} &
\includegraphics[width=4cm]{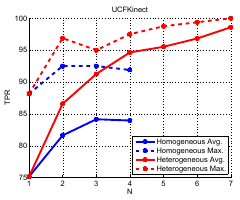} \\
(a) & (b)
\end{tabular}
\caption{TPR obtained using homogenous vs. heterogeneous labeled action instance sets for the MSRC-12(a) and the UCFKinect(b) datasets, as a function of set size $N$ (see text for
details). Due to limited number of action instances for each action-user combination in the UCFKinect dataset the homogeneous graphs terminate at $N = 4$.}
\label{SingleVsMultiTPR}
\end{center}
\end{figure}

We now present experimental validation of the intuitive
claims presented in Section~\ref{sec:HomoVsHeteroExplain}. In the following experiments,
a single action instance is randomly selected into
the training set for each user-action pair. A set of $N \geq 1$
randomly selected action instances is used for classifying
the user's identity at test time. In this experiment we assume
that the instances' action labels are given. Figs.~\ref{SingleVsMultiTPR}(a)
and~\ref{SingleVsMultiTPR}(b) show the average and the maximal TPR as a function
of $N$ using both homogeneous and heterogeneous sets,
for the MSRC-12 and the UCFKinect datasets, respectively.
For the homogeneous case, the ``Avg.'' graph shows the TPR
for randomly constructed homogeneous sets of size $N$, averaged
across all action types. The upper bound ``Max.''
graph shows the TPR for sets including instances from
the topmost, in terms of TPR, action type, for each $N$. 
For the heterogeneous case, the ``Avg.'' graph shows
the average TPR for sets of size $N$, containing a random
selection of action types. The ``Max.'' graph shows the TPR
for heterogeneous sets constructed of $N$ topmost, in terms
of TPR, action types. For both datasets, an average heterogeneous
case outperforms the homogeneous one, and even
outperforms the homogeneous' upper bound starting with
$N = 3$ for MSRC-12, and $N = 4$ for UCFKinect. As expected,
the selection of the best performing action types for
the heterogeneous set results in the best performance.

\subsection{Interactive Scenario}
\label{sec:res_cued_id}
We evaluate the cued identification approach on two datasets -- our own CuedId dataset and a publicly available MSRC-12 dataset.
Our experimental protocol contains two phases -- training and testing. 
Thus, we partition the response instances into disjoint training and test sets.
The training set contains, for each user, several response instances for each cue.
In training, we learn $p(\mathbf{x}|a,u)$ for $\left(a,u\right)\in\mathcal{A}\times\mathcal{U}$, and $p(a|c,u)$ for $\left(c,u\right)\in\mathcal{C}\times\mathcal{U}$.
The test phase is the actual interactive scenario where, given the selected cue, a relevant response instance is randomly selected from the test set.


\subsubsection{Skeleton Structure Normalization}
\label{sec:skel_norm}
Skeletal action instances combine two sources of features: structural and dynamic. Structural features, \eg, body height or length of limbs, 
represent individual anthropometric characteristics and may be used to infer personal attributes, such as gender or age, and identity. 
Roughly speaking, structural information comprises body part lengths and is action independent under a body part rigidity assumption. On the other hand, dynamic information encodes the way individuals move, and therefore, is action dependent. 
For example, Troje~\cite{trojePCA} shows that the dynamic features contain more information regarding the individual's gender than structural ones.
Because our focus in this work is on body motion, we use only dynamic information to represent individual motion patterns. There is an additional benefit in stripping the structural information from the skeleton. Since in our experiments we use a limited number of individuals, structural information may be deceptively useful for accurate identification. For example, using roughly the height of the skeleton may yield quite decent results. Nevertheless, increasing the number of subjects in the dataset by an order of magnitude is expected to significantly reduce performance. Thus, to eliminate such dataset bias, we apply a spatial normalization procedure~\cite{MovingPose} on each action instance by scaling all body part lengths to a predefined uniform template size.  

\subsubsection{Data Augmentation}
\label{sec:augmentation}
The application of the temporal and spatial normalization procedures gives us an additional benefit.
As said before, the number of individuals in existing datasets is quite limited -- on the order of magnitude of tens. 
However, given that action instances are normalized, it is possible to generate ``virtual'' individuals by synthesizing their motion patterns
using existing ones. 

Let $\mathbf{x}_1\in\mathbb{R}^d$ and $\mathbf{x}_2\in\mathbb{R}^d$ denote two spatially and temporally normalized instances of type $a$ performed by two different individuals $u_1$ and $u_2$ as a response to cue $c$.
It is possible to define a new action instance as a convex combination of $\mathbf{x}_1$ and $\mathbf{x}_2$, $\mathbf{x}_{1,2}^{\alpha} = \alpha\mathbf{x}_1+(1-\alpha)\mathbf{x}_2$, and to associate it with response $a$ to cue $c$, performed by some imaginary individual $u_{1,2}$, whose $a$ response to cue $c$ is $\alpha$-similar to that of $u_1$ and $(1-\alpha)$-similar to that of $u_2$. Of course, such instance generation cannot replace motion data obtained from real individuals, but it may approximate it in some sense. 
In our experiments we make use of such synthetic data augmentation to increase the number of individuals in the dataset. 
To create a virtual individual we select two real individuals $u_1,u_2\in\mathcal{U}$ and for each pair $(c,a)\in\mathcal{C}\times\mathcal{A}$ randomly choose $N$ response instance pairs,
$\left\{\left(\mathbf{x}_i^{u_1,c,a},\mathbf{x}_i^{u_2,c,a}\right)\right\}_{i=1}^N$. 
Each pair generates a new action instance of response $a$ to cue $c$ of the virtual individual $u_{1,2}$. 
In our experiments we use $\alpha \sim \mathcal{U}\left(0.3,0.7\right)$.

\begin{figure*}[tb]
\begin{center}
\begin{tabular}{cccc}
\includegraphics[width=4cm]{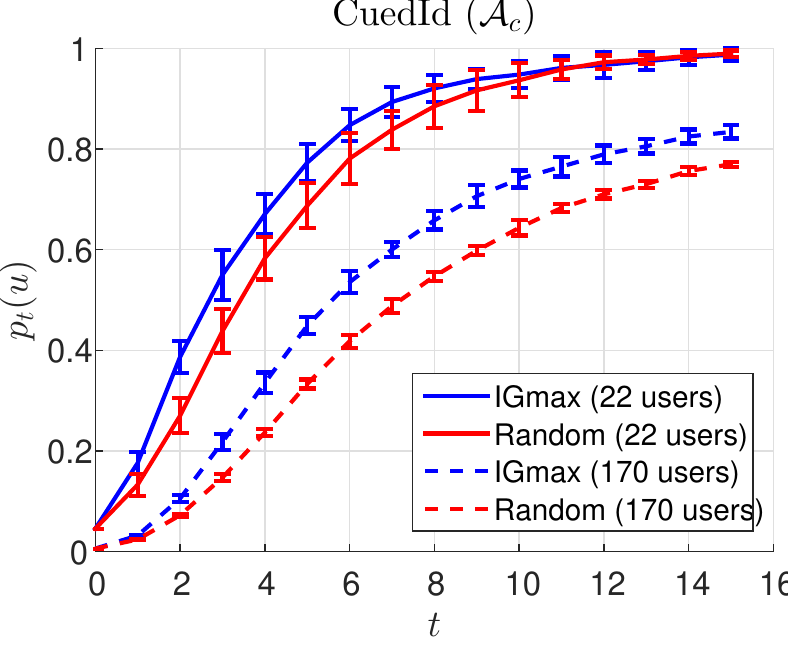} &
\includegraphics[width=4cm]{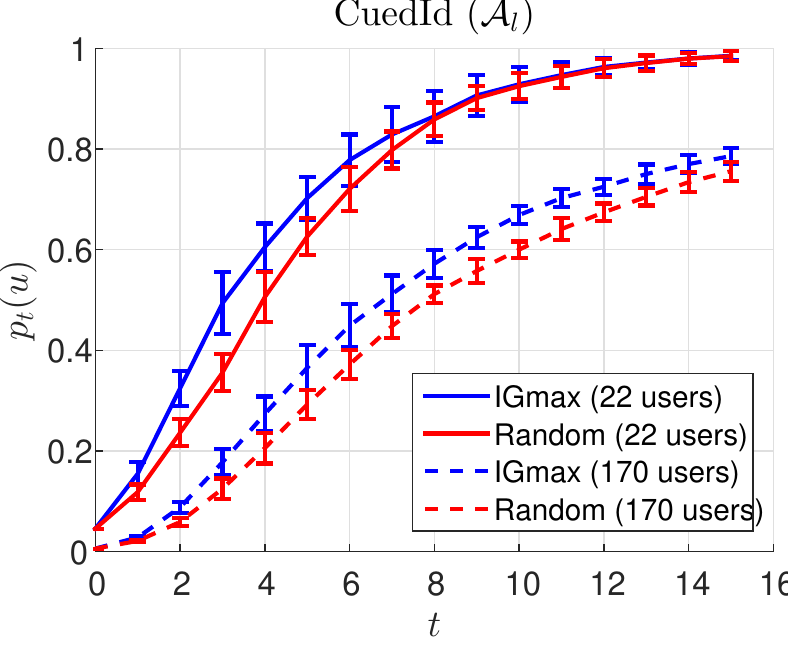} & 
\includegraphics[width=4cm]{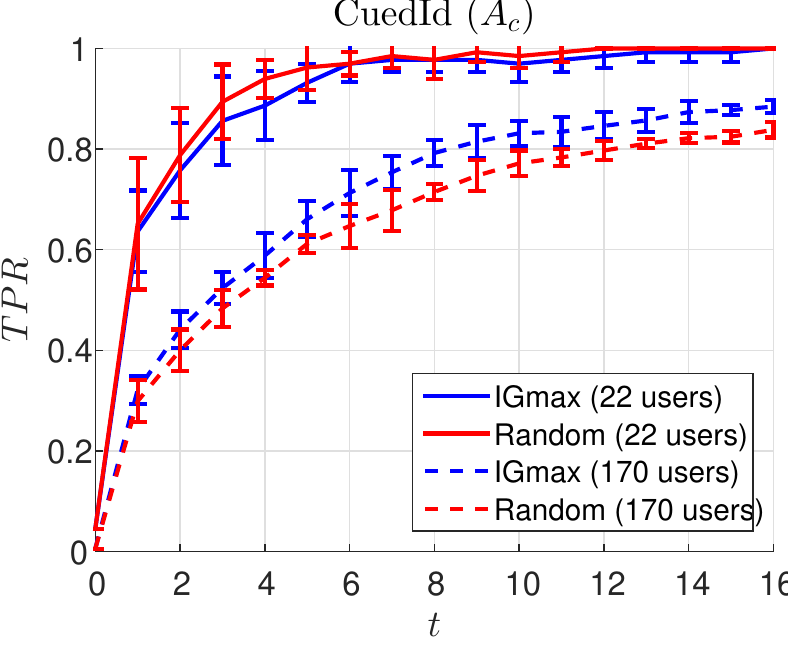} &
 \includegraphics[width=4cm]{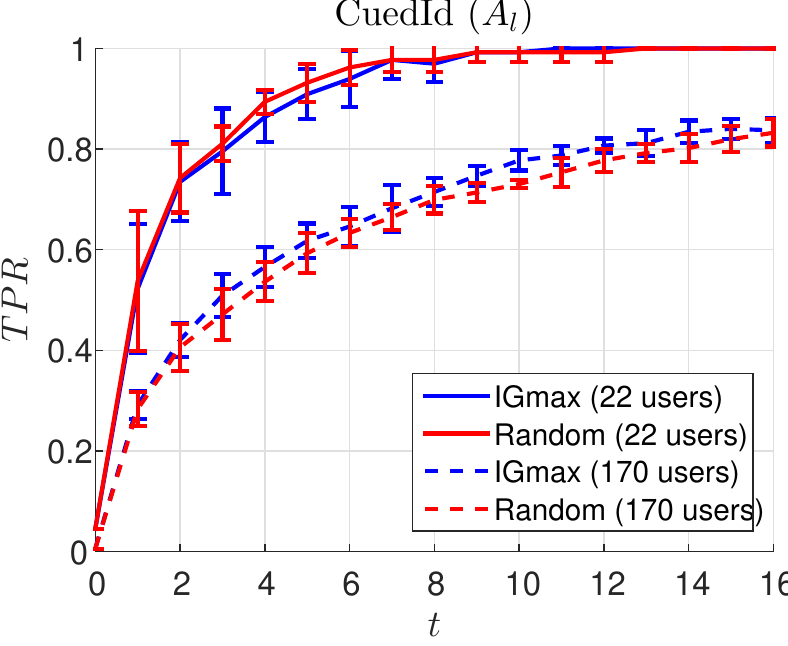} \\
(a) & (b) & (c) & (d)
\end{tabular}
\caption{(a-b) Performance of the CPI algorithm on the CuedId dataset for $\mathcal{A}_c$(a) and $\mathcal{A}_l$(b). The true class probability is displayed as a function of $t$. 
(c-d) Performance of the CPI algorithm on the CuedId dataset for $\mathcal{A}_c$(a) and $\mathcal{A}_l$(b). The true positive rte (TPR) is displayed as a function of $t$.
}
\label{fig:IG-MaxVsRandom-Cued}
\end{center}
\end{figure*}

\subsubsection{Experiments}


\textbf{CuedId Dataset.}
Seven response instances of each user--cue combination are used to learn the response likelihood probabilities during training.
Each subject is tested ten times and the results are averaged. 
The CPI algorithm is evaluated using two response type sets, $\mathcal{A}_c$ and $\mathcal{A}_l$, defined in Section~\ref{sec:ResTypeRep}.
We compare IG-Max selection criterion, defined in Section~\ref{sec:active}, to the baseline, Random, criterion.
To evaluate the CPI performance on larger numbers of subjects, we performed data augmentation as described in Section~\ref{sec:augmentation}.
Thus, we report results on two datasets -- the original one, containing 22 real subjects, and the augmented one, containing additional 148 synthetic subjects, 170 in total.

Figs.~\ref{fig:IG-MaxVsRandom-Cued}(a,b) demonstrate the results in terms of true class probability as a function of the iteration index, $t$, for both variants of $\mathcal{A}$.
In all experiments, IG-Max obtains statistically significant better performance than Random. 
For example, it takes an average of seven IG-Max selections to identify any subject from the original dataset, with 90\% confidence.
However, about nine Random selections are required to reach the same level of confidence. As expected, this gap widens as the number of enrolled subjects grows.
Note that the CPI performance degrades gracefully with an increase in the number of subject in the system. 
For example, given 22 enrolled subjects, it takes an average of 4.5 iterations to reach a confidence level of 80\%. 
With 170 subjects, the same confidence level is reached in 13 iterations on average.

Comparing two choices for $\mathcal{A}$, we see that $\mathcal{A}_l$ results in a slightly worse performance than $\mathcal{A}_c$.
However, note that the loss in accuracy comes with a significant benefit -- reduction in the generative model's complexity. 
Recall that, based on our discussion in Section~\ref{sec:ResTypeRep}, it is more efficient and scalable to model $p(\mathbf{x}|a,u)$ and $p(a|c,u)$ than 
to model $p(\mathbf{x}|c,u)$ directly.

\textbf{MSRC-12 Dataset.}
To adapt the dataset to an interactive scenario, we assume that subjects
performed the gestures as a response to a visual cue, and that there is a one-to-one correspondence between the cue and the response: 
$\mathcal{C}=\mathcal{A}=\{1,\ldots,12\}$.

\begin{figure}[ht]
\begin{center}
\begin{tabular}{cc}
\includegraphics[width=4cm]{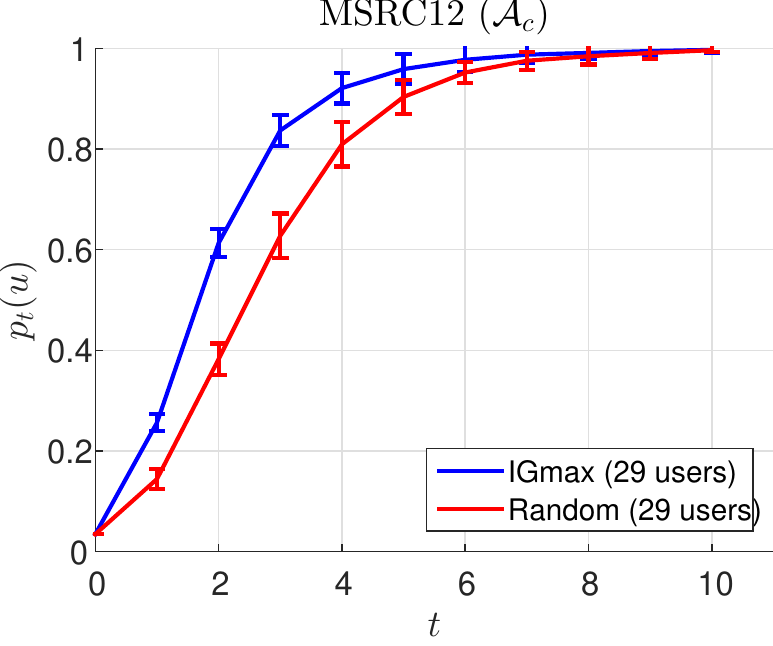} &
\includegraphics[width=4cm]{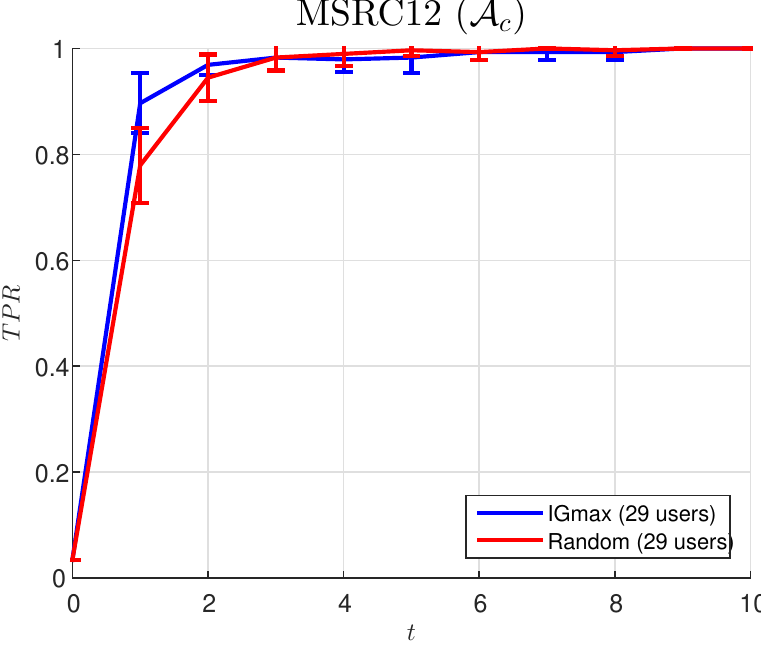} \\
(a) & (b)
\end{tabular}
\caption{Performance of the CPI algorithm on the MSRC-12 dataset. 
The true class probability (a) and the true positive rate (b) are displayed as a function of $t$.
}
\label{fig:IG-MaxVsRandom-MSRC12}
\end{center}
\end{figure}

In training we use four instances for each user--cue combination. and the rest for testing.
Each of the 29 individuals is tested 10 times and the results are averaged.
Figs.~\ref{fig:IG-MaxVsRandom-MSRC12}(a) and~\ref{fig:IG-MaxVsRandom-MSRC12}(b)  shows the results in terms of the average true class probability and true positive rate, as a function of $t$, respectively.
Although, there are more subjects than in the CuedId dataset, the performance is better. An average confidence level of 82\% is reached
after just three cues. Although IG-Max's TPR is not significantly higher than Random's, its uncertainty is significantly lower. Also, the gap between the IG-Max and Random is more significant than in the CudeId dataset experiments. 
This is probably because unlike the CuedId, gestures in MSRC-12 have a semantic meaning, and therefore are more discriminative.

\subsubsection{Open Set Recognition}
Up until now, our approach was evaluated under the closed-set assumption, namely, we assumed that each individual in the test set belongs to a group of already enrolled identities. Speaking of biometric applications, a natural question is what happens when we reveal this assumption.    
In this section we evaluate our approach in the \emph{open-set} setting. Under this setting, the query identity may or may not belong to the 
users atomic set. The performance of the entire system is then combined of two evaluation metrics. The first one is the accuracy in determining whether the individual belongs to the atomic set of enrolled identities, and the second one is the accuracy under the close-set assumption. 

\begin{figure*}[tb]
\begin{center}
\begin{tabular}{cccc}
\includegraphics[width=4cm]{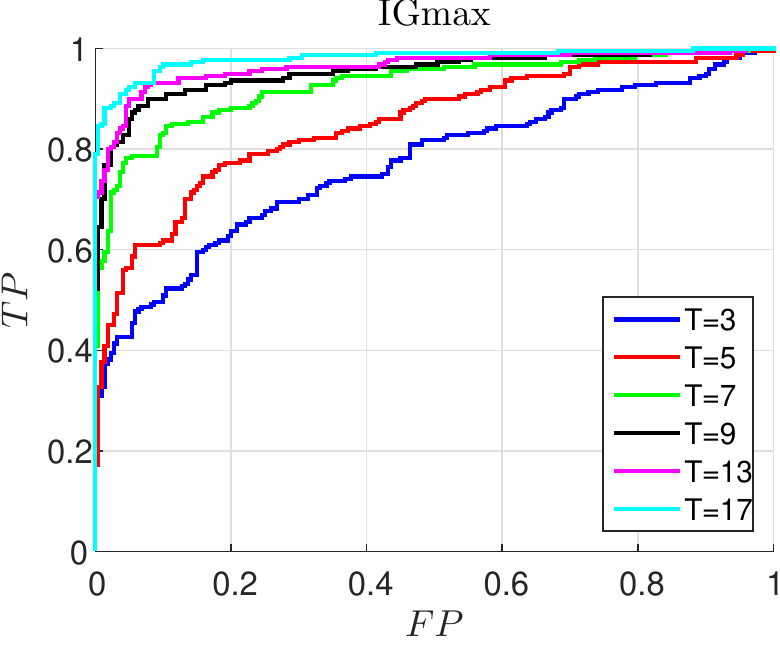} &
\includegraphics[width=4cm]{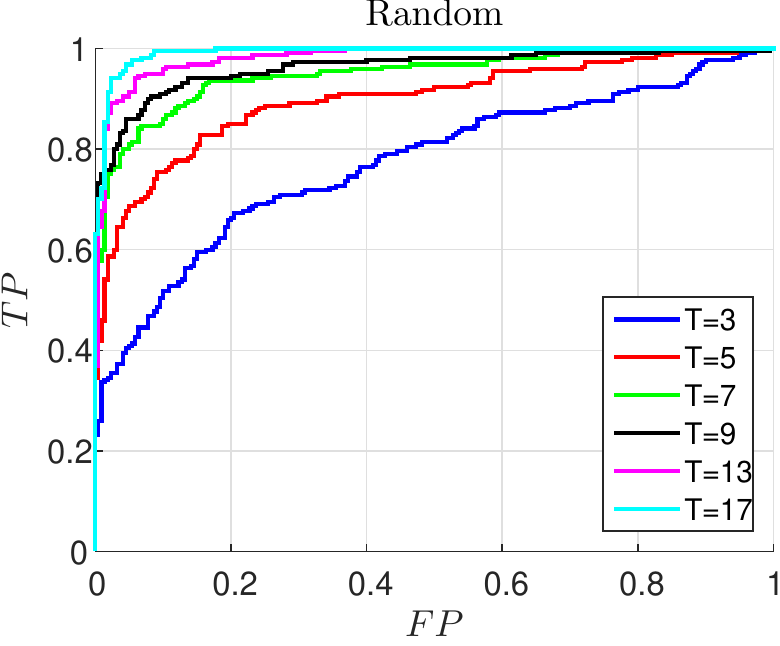} & 
\includegraphics[width=4cm]{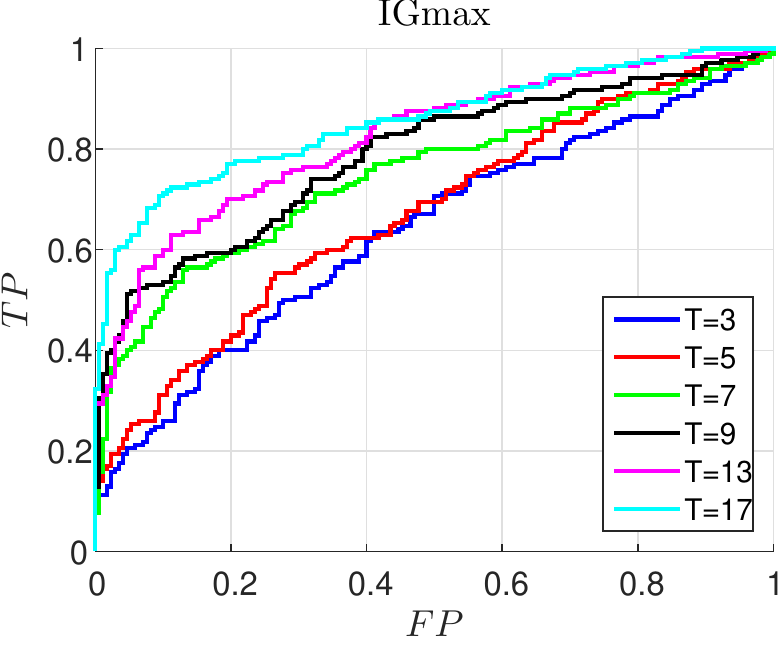} &
 \includegraphics[width=4cm]{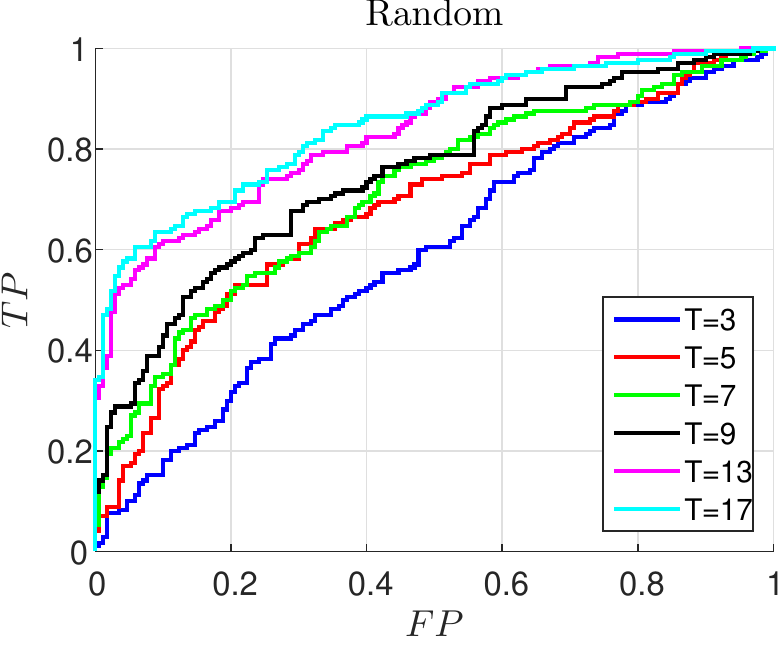} \\
(a) & (b) & (c) & (d)
\end{tabular}
\caption{ROC curves describing the performance of the CPI algorithm under the open-set setting. 
(a,b) 22 identities. (c,d) 170 identities. 
}
\label{fig:ROC}
\end{center}
\end{figure*}

We pose the problem of classifying whether or not the user was previously enrolled, as a binary classification task.   
To this end, we use the MAP probability output of the CPI algorithm as a detection threshold. Fig.~\ref{fig:ROC}(a,b) and Fig.~\ref{fig:ROC}(c,d) show the ROC curves for the 22 and 170 user atomic sets, respectively, for both the Random and the IGmax cue selection strategies under various budget constraints ($T$). 
As expected the performance improves with $T$, as the algorithm makes its decision based on more user samples. 
As with the closed-set setting reported in Fig.~\ref{fig:IG-MaxVsRandom-Cued}, the Random strategy does better for the 22 identities, however as more identities are added, it is outperformed by the IGmax strategy. 

\section{Discussion and Conclusions}
\label{sec:disc}
In this work we introduce a general framework for person identification under three scenarios using contextual motion only. For the Login and Smart home scenarios, our experiments on publicly available action recognition and person authentication datasets, demonstrate that the approach's accuracy is high enough to be practical. As shown by our experiments, deep learning approaches result in a better accuracy as the number of identities grow. We hypothesize that combining our generative models with the power of deep learning is expected to improve the results even further. For example, more powerful action representation learned using deep models can potentially improve the performance of each one of the considered scenarios.

For the novel Interactive scenario, we propose an active solution, based on an information theory. We validated the approach on our own CuedId dataset collected using a specially designed system, and on a publicly available MSRC-12 dataset.
The results reported in this work are an initial step in a new direction that opens avenues for future research. 
While here we have assumed that the set of cues is discrete and predefined, it would be beneficial 
to define the cue as a random variable over a continuous domain of its parameters, allowing us to generalize to cues never generated before.
Even though we have chosen the problem of person identification as a test case, the approach may be also used for classification of personal attributes such as age and gender, or estimation of medical conditions such as Parkinson's disease. 

\ifCLASSOPTIONcaptionsoff
  \newpage
\fi



%

\bibliographystyle{abbrv}
\bibliography{CuedIdPAMI}

\vspace{4pt}\noindent\textbf{Igor Kviatkovsky} received the BSc, MSc and PhD degrees in computer science from the Technion-Israel Institute of Technology. 
He is currently a Senior Research Scientist at the Amazon Go research team. 
His research interests include computer vision, machine learning and computer graphics.

\vspace{4pt}\noindent\textbf{Ehud Rivlin} received the BSc and MSc degrees
in computer science and the MBA degree from
the Hebrew University in Jerusalem and the PhD
from the University of Maryland. Currently, he is
a professor in the Computer Science
Department at the Technion-Israel Institute of
Technology. His current research interests are in
machine vision and robot navigation.

\vspace{4pt}\noindent\textbf{Ilan Shimshoni} received the BSc degree in
mathematics and computer science from the
Hebrew University Jerusalem, Israel, the MSc
degree in computer science from the Weizmann
Institute of Science, Rehovot, Israel, and the PhD
degree in computer science from the University
of Illinois at Urbana-Champaign. Currently, he is
a professor at the Department of Information Systems,
Haifa University. His current research interests
include computer vision, robotics, and
computer graphics. He is a member of the IEEE.

\pagebreak

\appendices
\section{}
Figs.~\ref{fig:GM}(a-c) show the graphical probabilistic models corresponding to the three scenarios introduced in Section~\ref{sec:GenModel}. 
Each model forms a directed acyclic graph (DAG). It is a well known fact~\cite{bishop} that the joint distribution of all the random variables 
modeled by the graph, is given by the product, over all of the nodes of the graph, of a conditional distribution for each node, conditioned on the variables corresponding to the parents of that node in the graph.
Thus, the joint probability of all the variables in models in Figs.~\ref{fig:GM}(a,b) and Fig.~\ref{fig:GM}(c) are:
\begin{equation}
\label{eq:jointProbLogin}
p(u,\left\{\mathbf{x}_i,a_i\right\}_{i=1}^N)=p(u)\prod_{i=1}^N p(\mathbf{x}_i|a_i,u)p(a_i),
\end{equation}

and,
\label{eq:jointProbCuedId}
\begin{align*}
p(u,\left\{\mathbf{x}_i,a_i,c_i\right\}_{i=1}^N) = p(u)\prod_{i=1}^N p(\mathbf{x}_i|a_i,u)p(a_i|c_i,u)p(c_i),
\end{align*}

respectively.

Note that, $a_i\bigCI a_j \mid \emptyset$, for $i\neq j$, due to the D-separation property~\cite{pearl}, and therefore:
\begin{equation}
\label{eq:aIndep}
p(\left\{a_i\right\}_{i=1}^N)=\prod_{i=1}^N p(a_i).
\end{equation}

\subsection{Login Scenario}
\label{appendix:ProbLogin}
Given a set of $N$ labeled action instances, $\left\{\mathbf{x}_i,a_i|i=1,\ldots N, \mathbf{x}_i\in\mathbb{R}^d, a_i\in\mathcal{A}\right\}$, the conditional probability of the user identity given the labeled instances is, 
\begin{align}
\nonumber &p(u|\left\{\mathbf{x}_i,a_i\right\}_{i=1}^N) 
= \frac{p(u,\left\{\mathbf{x}_i,a_i\right\}_{i=1}^N)}{p(\left\{\mathbf{x}_i,a_i\right\}_{i=1}^N)} \\
\nonumber &\stackrel{Eq.~\ref{eq:jointProbLogin}}{=} \frac{p(u)\prod_{i=1}^N
p(\mathbf{x}_i|a_i,u)p(a_i)}{p(\left\{\mathbf{x}_i\right\}_{i=1}^N|\left\{a_i\right\}_{i=1}^N)p(\left\{a_i\right\}_{i=1}^N)} \\
\nonumber &\stackrel{Eq.~\ref{eq:aIndep}}{=}\frac{p(u)\prod_{i=1}^N p(\mathbf{x}_i|a_i,u)}{p(\left\{\mathbf{x}_i\right\}_{i=1}^N|\left\{a_i\right\}_{i=1}^N)} \\
&\propto p(u)\prod_{i=1}^N p(\mathbf{x}_i|a_i,u).
\end{align}

\subsection{Person Identification in Smart Home Scenario}
\label{appendix:ProbSmartHome}
Given a set of $N$ action instances, $\left\{\mathbf{x}_i|i=1,\ldots N, \mathbf{x}_i\in\mathbb{R}^d \right\}$, the conditional probability of the user identity given the instances is, 
\begin{align}
\nonumber &p(u|\left\{\mathbf{x}_i\right\}_{i=1}^N)
=\frac{p(u,\left\{\mathbf{x}_i\right\}_{i=1}^N)}{p(\left\{\mathbf{x}_i\right\}_{i=1}^N)} \\
\nonumber &=\frac{\sum_{\left\{a_i\right\}_{i=1}^N \in\mathcal{A}^N}p(u,\left\{\mathbf{x}_i,a_i\right\}_{i=1}^N)}{p(\left\{\mathbf{x}_i\right\}_{i=1}^N)} \\
\nonumber &\stackrel{Eq.~\ref{eq:jointProbLogin}}{=}\frac{p(u)\sum_{\left\{a_i\right\}_{i=1}^N \in\mathcal{A}^N}\prod_{i=1}^N
p(\mathbf{x}_i|a_i,u)p(a_i)}{p(\left\{\mathbf{x}_i\right\}_{i=1}^N)} \\
\nonumber &=\frac{p(u)\prod_{i=1}^N\sum_{a_i\in\mathcal{A}}
p(\mathbf{x}_i|a_i,u)p(a_i)}{p(\left\{\mathbf{x}_i\right\}_{i=1}^N)} \\
\nonumber &=\left(\frac{\prod_{i=1}^N p(\mathbf{x}_i)}{p(\left\{\mathbf{x}_i\right\}_{i=1}^N)}\right)
p(u)\prod_{i=1}^N\sum_{a_i\in\mathcal{A}}
\frac{p(\mathbf{x}_i|a_i,u)}{p(\mathbf{x}_i|a_i)}p(a_i|\mathbf{x}_i) \\
&\propto p(u)\prod_{i=1}^N \sum_{a_i\in\mathcal{A}}\frac{p(\mathbf{x}_i|a_i,u)}{p(\mathbf{x}_i|a_i)}p(a_i|\mathbf{x}_i),
\end{align}
where we have marginalized over all possible labeling combinations of $N$ instances, 
$\left\{a_i \right\}_{i=1}^N \in \mathcal{A}^N$.

\subsection{Interactive Person Identification}
\label{appendix:ProbCuedId}
First, note that, $c_i \bigCI c_j \mid \emptyset$, for $i\neq j$, due to the D-separation property, and therefore:
\begin{equation}
\label{eq:cIndep}
p(\left\{c_i\right\}_{i=1}^N)=\prod_{i=1}^N p(c_i).
\end{equation}

Here we assume that we are given a set of $N$ action instances, $\left\{\mathbf{x}_i|i=1,\ldots N, \mathbf{x}_i\in\mathbb{R}^d \right\}$, generated as a response to a set of cues, 
$\left\{c_i|i=1,\ldots N, c_i\in\mathcal{C}\right\}$. The response type is introduced via latent variables $\left\{a_i|i=1,\ldots N, a_i\in\mathcal{A}\right\}$ coming from a set of possible actions, $\mathcal{A}$. The probability of user identity given the instances obtained as a response to the cues is, 
\begin{align}
\nonumber &p(u|\left\{\mathbf{x}_i,c_i\right\}_{i=1}^N) 
=\frac{p(u,\left\{\mathbf{x}_i,c_i\right\}_{i=1}^N)}{p(\left\{\mathbf{x}_i,c_i\right\}_{i=1}^N)} \\
\nonumber &=\frac{\sum_{\left\{a_i\right\}_{i=1}^N \in\mathcal{A}^N}p(u,\left\{\mathbf{x}_i,a_i,c_i\right\}_{i=1}^N)}{p(\left\{\mathbf{x}_i,c_i\right\}_{i=1}^N)} \\
\nonumber &\stackrel{Eq.~\ref{eq:jointProbCuedId}}{=}\frac{\sum_{\left\{a_i\right\}_{i=1}^N \in\mathcal{A}^N}p(u)\prod_{i=1}^N
p(\mathbf{x}_i|a_i,u)p(a_i|c_i,u)p(c_i)}{p(\left\{\mathbf{x}_i,c_i\right\}_{i=1}^N)} \\
\nonumber &=\frac{p(u)\prod_{i=1}^N p(c_i)\sum_{a_i\in\mathcal{A}}
p(\mathbf{x}_i|a_i,u)p(a_i|c_i,u)}{p(\left\{\mathbf{x}_i\right\}_{i=1}^N|\left\{c_i\right\}_{i=1}^N)p(\left\{c_i\right\}_{i=1}^N)} \\
\nonumber &\stackrel{Eq.~\ref{eq:cIndep}}{=}\frac{p(u)\prod_{i=1}^N\sum_{a_i\in\mathcal{A}}
p(\mathbf{x}_i|a_i,u)p(a_i|c_i,u)}{p(\left\{\mathbf{x}_i\right\}_{i=1}^N|\left\{c_i\right\}_{i=1}^N)} \\
&\propto p(u)\prod_{i=1}^N \sum_{a_i\in\mathcal{A}}p(\mathbf{x}_i|a_i,u)p(a_i|c_i,u).
\end{align}

\subsection{Active Cue Selection}
\label{appendix:IG}
From the generative model in Fig.~\ref{fig:GM}(d) (we omit the subscript $t$ for clarity) and the D-separation property, it follows that $u \bigCI c \mid \emptyset$, and thus,
\begin{equation}
\label{eq:ucIndep}
p(u|c)=p(u).
\end{equation}

Now, we use basic information theory axioms~\cite{CoverThomas} to express the mutual information of the user and the response instance given the cue, 
\begin{align}
\nonumber &I(u;\mathbf{x}|c)
\nonumber = H(u|c) - H(u|\mathbf{x},c) \\
\nonumber &= H(u|c) - H(u,\mathbf{x}|c) + H(\mathbf{x}|c) \\
\nonumber &= -\sum_{u\in\mathcal{U}}p(u|c)\log p(u|c) +\sum_{u\in\mathcal{U}}\int_{\mathcal{X}} p(u,\mathbf{x}|c)\log p(u,\mathbf{x}|c)d\mathbf{x} -\\ 
\nonumber &~~~~~~~~~\int_{\mathcal{X}}p(\mathbf{x}|c)\log p(\mathbf{x}|c)d\mathbf{x} \\
\nonumber &= \sum_{u\in\mathcal{U}}\int_{\mathcal{X}} p(u,\mathbf{x}|c)\log p(u,\mathbf{x}|c)- p(u,\mathbf{x}|c)\log p(u|c)-\\
\nonumber &~~~~~~~~~~~~~~~~~p(u,\mathbf{x}|c)\log p(\mathbf{x}|c)d\mathbf{x} \\
\nonumber &= \sum_{u\in\mathcal{U}}\int_{\mathcal{X}}  p(u,\mathbf{x}|c)\log\frac{p(u,\mathbf{x}|c)}{p(\mathbf{x}|c)p(u|c)}d\mathbf{x} \\
\nonumber &=\sum_{u\in\mathcal{U}}p(u|c)\int_{\mathcal{X}}  p(\mathbf{x}|c,u)\log\frac{p(\mathbf{x}|c,u)p(u|c)}{p(\mathbf{x}|c)p(u|c)}d\mathbf{x} \\
&\stackrel{Eq.~\ref{eq:ucIndep}}{=}\sum_{u\in\mathcal{U}}p(u)\int_{\mathcal{X}}  p(\mathbf{x}|c,u)\log\frac{p(\mathbf{x}|c,u)}{p(\mathbf{x}|c)}d\mathbf{x}.
\end{align}

\vfill




\end{document}